\pdfoutput=1

\documentclass[11pt]{article}

\usepackage{EMNLP2022}

\usepackage{array}
\usepackage{times}
\usepackage{latexsym}
\usepackage[T1]{fontenc}

\usepackage[utf8]{inputenc}
\usepackage{booktabs}
\usepackage{microtype}

\newcommand{\eat}[1]{}

\title{\ours: Memory-assisted Prompt Editing with User Feedback}

\author{Aman Madaan~\thanks{\hspace{0.5em}Equal Contribution}\hspace{0.5em}, Niket Tandon~\footnotemark[1]\hspace{0.5em}$^\dagger$,  Peter Clark$^\dagger$, Yiming Yang \\
  Language Technologies Institute, Carnegie Mellon University, Pittsburgh, PA, USA \\
  $^\dagger$ Allen Institute for Artificial Intelligence, Seattle, WA, USA \\ 
  \texttt{\{amadaan,yiming\}@cs.cmu.edu} \\ \texttt{\{nikett,peterc\}@allenai.org} \\} 
  
\usepackage{xspace}
\usepackage{graphicx} 
\usepackage{subcaption} 
\usepackage{soul}
\usepackage{pifont} 
\usepackage{listings}
\usepackage{amsmath}

\DeclareMathOperator*{\argmin}{arg\,min}

\definecolor{cosmiclatte}{rgb}{1.0, 0.97, 0.91}
\definecolor{codegreen}{rgb}{0,0.6,0}
\definecolor{codegray}{rgb}{0.5,0.5,0.5}
\definecolor{codepurple}{rgb}{0.58,0,0.82}
\definecolor{backcolour}{rgb}{0.95,0.95,0.92}

\lstdefinestyle{mystyle}{
    backgroundcolor=\color{backcolour},   
    commentstyle=\color{codegreen},
    keywordstyle=\color{magenta},
    numberstyle=\tiny\color{codegray},
    stringstyle=\color{codepurple},
    basicstyle=\ttfamily\footnotesize,
    breakatwhitespace=false,         
    breaklines=true,                 
    captionpos=b,                    
    keepspaces=true,                 
    numbers=left,                    
    numbersep=5pt,                  
    showspaces=false,                
    showstringspaces=false,
    showtabs=false,                  
    tabsize=2
}

\usepackage{pgfplotstable}

\definecolor{Red}{rgb}{1,0,0}
\definecolor{Green}{rgb}{0.4,1,0.2}
\definecolor{Blue}{rgb}{0,0,1}
\definecolor{Red}{rgb}{0.9,0,0}
\definecolor{Orange}{rgb}{1,0.5,0}
\definecolor{yellow}{rgb}{0.65,0.6,0}
\definecolor{cadmiumgreen}{rgb}{0.2, 0.7, 0.24}
\definecolor{verbcolor}{HTML}{13B584}
\newcommand{\V}[1]{\mathbf{#1}}

\newcommand{\verbalization}[1]{\textcolor{verbcolor}{#1}}

\newcommand{\emnlpcr}[1]{#1}

\newcommand{\ourir}{\textsc{gud-ir}\xspace}
\newcommand{\user}{\textcolor{blue}{User:}\xspace}

\newcommand{\csrr}[1]{\textcolor{black}{#1}}

\newcommand{\csrrcr}[1]{\textcolor{black}{#1}}

\newcommand{\vtwo}[1]{{#1}}

\newcommand{\secref}[1]{\S\ref{#1}}

\newcommand{\gpt}{\textsc{gpt-3-175b}\xspace} 
\newcommand{\kate}{\textsc{kate}\xspace}
\newcommand{\webqa}{\textsc{webqa}\xspace} 
\newcommand{\gptshort}{\textsc{gpt-3}\xspace} 
 
\newcommand{\ours}{MemPrompt\xspace}

\newcommand{\delphi}{\textsc{delphi}\xspace}
\newcommand{\nl}{\textsc{nl}\xspace}

\newcommand{\instr}{\textsc{ins}\xspace}

\newcommand{\good}{\textsc{good}\xspace}
\newcommand{\bad}{\textsc{bad}\xspace}
\newcommand{\okay}{\textsc{okay}\xspace}

\newcommand{\ert}{\textsc{ert}\xspace}
\newcommand{\ertnl}{\textsc{ert-nl}\xspace}
\newcommand{\ertcat}{\textsc{ert-cat}\xspace}

\newcommand{\cat}{\textsc{cat}\xspace}
\newcommand{\ques}{\V{x}}
\newcommand{\ans}{\V{y}}
\newcommand{\ra}{\V{u}}
\newcommand{\fb}{\mathbf{fb}}

\newcommand{\sep}{\#}
\newcommand{\prompt}{\V{p}}
\newcommand{\memory}{\mathcal{M}}
\newcommand{\syn}{syn\xspace}
\newcommand{\ant}{ant\xspace}
\newcommand{\defn}{defn\xspace}
\newcommand{\sent}{sent\xspace}

\newcommand{\qa}{\textsc{qa}\xspace}
\newcommand{\homn}{hom\xspace}

\newenvironment{des}{                 
     \parskip 0cm \begin{list}{}{\parsep 0cm \itemsep 0cm \topsep 0cm}}{
       \end{list}} 

\newcommand{\quesm}{$\ques$\xspace}
\newcommand{\ansm}{$\ans$\xspace}
\newcommand{\ram}{$\ra$\xspace}
\newcommand{\fbm}{$\V{fb}$\xspace}

\newcommand{\fbsample}{$(\ques, \fb \rightarrow \ra , \ans)$\xspace}
\newcommand{\fprobi}{$Pr(\V{fb}_i)$\xspace}
\newcommand{\memorym}{$\memory$\xspace}

\newcommand{\retm}{$\memory(\ques)$\xspace}

\newcommand{\promptm}{$\prompt$\xspace}

\newcommand{\ie}{i.e.,\xspace}
\newcommand{\eg}{e.g.,\xspace}

\newcommand{\nomem}{\textsc{no-mem}\xspace}
\newcommand{\growprompt}{\textsc{grow-prompt}\xspace}

\newcommand\ABox[2]{
  \fbox{\lower0.75cm
    \vbox to 1.5cm{\vfil
      \hbox to 2.1cm{\hfil\parbox{2.9cm}{#1\\#2}\hfil}
      \vfil}%
  }%
}

\newcommand{\tf}{\texttt{T5}\xspace}

\newcommand{\sts}{\textsc{seq2seq}\xspace}

\def\@withdot.{\ifmmode\!\string/\!
               \else\kern-1.8pt\string/\kern-1.8pt\fi.}

\newcommand{\squishlist}{
  \begin{list}{$\bullet$}
    { \setlength{\itemsep}{0pt}      \setlength{\parsep}{3pt}
      \setlength{\topsep}{3pt}       \setlength{\partopsep}{0pt}
      \setlength{\leftmargin}{1.5em} \setlength{\labelwidth}{1em}
      \setlength{\labelsep}{0.5em} } }
\newcommand{\reallysquishlist}{
  \begin{list}{$\bullet$}
    { \setlength{\itemsep}{0pt}    \setlength{\parsep}{0pt}
      \setlength{\topsep}{0pt}     \setlength{\partopsep}{0pt}
      \setlength{\leftmargin}{0.2em} \setlength{\labelwidth}{0.2em}
      \setlength{\labelsep}{0.2em} } }

 \newcommand{\squishend}{
     \end{list} 
 }

\begin{document}
\maketitle
\begin{abstract}
Large LMs such as \gptshort are powerful, but can commit mistakes that are obvious to humans.
For example, \gptshort would mistakenly interpret "What word is similar to \textit{good}?" to mean a homophone, while the user intended a synonym. Our goal is to effectively correct such errors via user interactions with the system but without retraining, which will be prohibitively costly. We pair \gptshort with a growing memory of recorded cases where the model misunderstood the user's intents, along with user feedback for clarification.
Such a memory allows our system to produce enhanced prompts for any new query based on the user feedback for error correction on similar cases in the past.
On four tasks (two lexical tasks, two \csrr{advanced} ethical reasoning tasks), we show how a (simulated) user can interactively teach a deployed \gptshort, substantially increasing its accuracy over the queries with different kinds of misunderstandings by the \gptshort.
Our approach is a step towards the low-cost utility enhancement for very large pre-trained LMs.\footnote{Code, data, and instructions to implement \ours for a new task at \url{https://www.memprompt.com/}}


\end{abstract}





\section{Introduction}

\begin{figure}[!t]
\centerline{
\fbox{
    \parbox{0.49\textwidth}{
	\underline{Our memory enhanced \gptshort implementation.}
\begin{des}
\item[{\bf \user}] What word is similar to \textit{good}?
\item[{\bf \gptshort:}] The homophone of good is: wood.
\item[{\bf \user}] "Similar to" means "with similar meaning".
\item[{\bf \gptshort:}] Noted {\it [writes to memory]}
\item[{\bf \user}] What word is similar to \textit{surprised}?
\item[{\bf \gptshort:}] The synonym of surprised is: amazed. \\{\it [Retrieves and adds to prompt `"Similar to" means "with similar meaning"']}.
\end{des}
    }
}}
\caption{This paper enhances \gptshort performance by looking up questions with a similar intent that received any user feedback. Our approach is simple because only the \csrr{question in the prompt} needs to be updated with relevant feedback, and no retraining is necessary.}
\label{fig:running-example}
\end{figure}

\csrr{Language models are now better than ever before at generating realistic content, but still lack commonsense \cite{bender-koller-2020climbing,marcus_gpt3}. One failure mode due to a lack of commonsense is in misunderstanding a user's \textit{intent}. The typical remedy of retraining with more data is prohibitive due to the cost and infrastructure requirements. In such cases, even if users repeatedly observe the model making a mistake, there are no avenues to provide feedback to the model to make it more accurate and personalized over time.}

\csrr{Our goal is to allow users to correct such errors directly through interaction, and without retraining by injecting the knowledge required to correct the model's misunderstanding. 
Building upon the recent success of injecting commonsense in the input \citep{Lewis2020RetrievalAugmentedGF, talmor2020leapofthought}, we propose a novel approach of injecting knowledge in the input via interactive feedback from an end-user.}

\begin{figure*}[t]
\centering
    \includegraphics[scale=0.25]{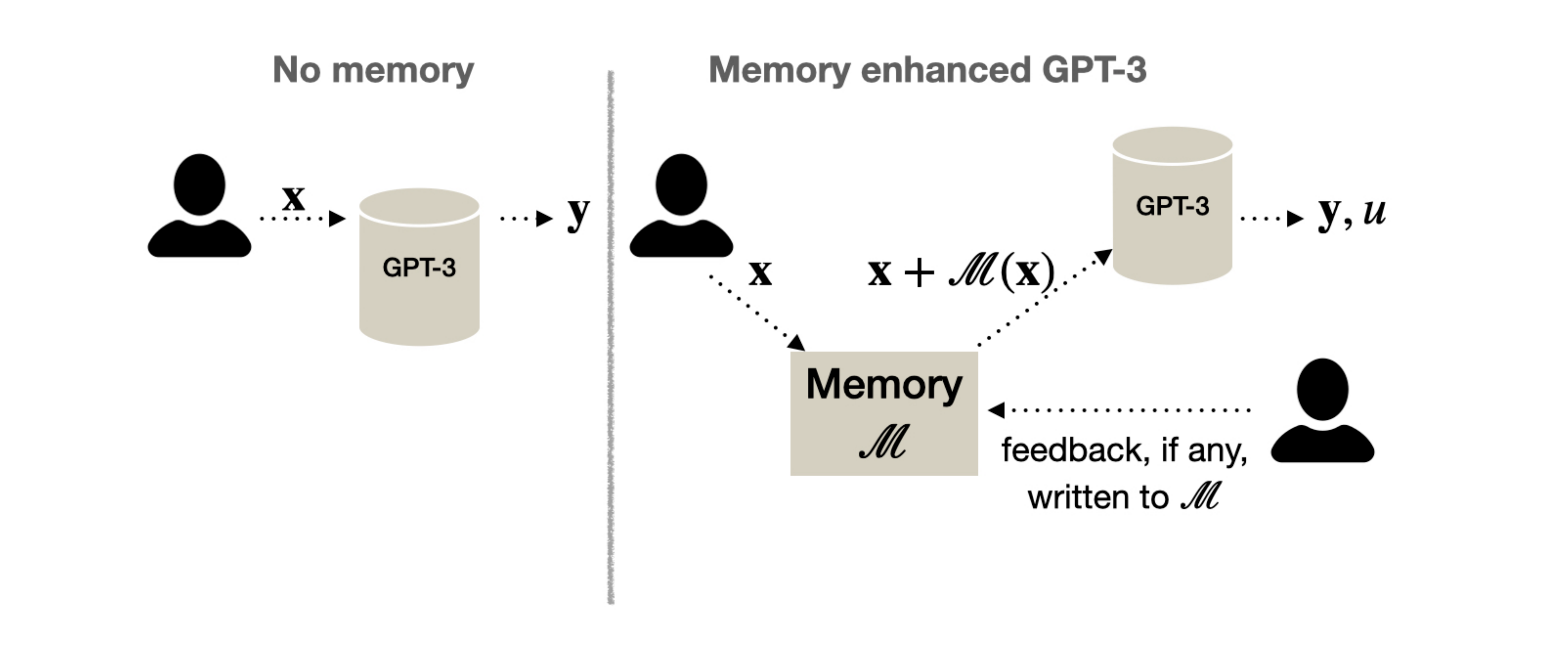}
    \caption{Proposed architecture: (left) \gptshort does not account for user feedback. (right) \ours maintains a memory $\memory$ 
          of corrective feedback, and searches for feedback from prior queries
    with a similar intent as $x$ using a retrieval function \retm. $x$ is then concatenated to the retrieved feedback and appended to the prompt for querying \gptshort. Users can also give new feedback on the model's task understanding $u$, then added to $\memory$.}
    \label{fig:method}
\end{figure*}


Our approach, \ours,  pairs \gptshort with a growing memory of cases where the model misunderstood user's intent and was provided with corrective feedback. 
This feedback is question dependent, and thus the prompt for each sample is \textit{edited} to adapt to the input.
In this sense, our work can be seen as an instance of prompt engineering~\cite{Liu2021PretrainPA} which involves editing the prompts. Our work adds interactivity to prompt engineering as it involves dynamically updating the prompt for every instance.

Figure \ref{fig:running-example} presents a sample interaction between a user and \gptshort that our setup enables.
The model was asked for a similar word. However, the model's (incorrect) task understanding \ram was ``The homophone of good is''.
The user can detect such discrepancy between the intended and interpreted task instruction, and can provide feedback $\fb$ as "\textit{similar to} means \textit{with a similar meaning}", clarifying that they actually wanted a synonym.
Crucially, note that such instructional correction is feasible {\it even if the user does not know
the correct answer to their question}, as they are critiquing the model's understanding of their
intent, rather than the answers themselves.
Thus, our setup \textbf{does not} require the users to be experts at tasks being solved, another advantage of our approach.

Further, it is desirable to have a system that can leverage past feedback on new, unseen examples for prompt-editing. We maintain a memory $\memory$ of such feedback as a set of key-value pairs, where the
key is a misunderstood question, and the value is the user's feedback to correct that misunderstanding. Given a new question, we check if the model has made a mistake
on a similar question earlier, by querying the memory for a similar question. If found,
append the corresponding feedback to the question prompt. This mechanism aims to
prevent the model from making the same type of mistake twice. This failure-driven reminding
mechanism draws inspiration from the theory of recursive reminding in psychology \cite{Jacoby2013},
which suggests humans index error corrections in the context in which those errors occurred.

This paper presents the general architecture for the system and provides representative implementations for each component.
We then demonstrate the system on four tasks, using simulated user feedback:
(1) lexical relations (e.g., antonyms, Figure \ref{fig:running-example}),
(2) word scrambling (e.g., anagrams), (3) ethical reasoning with user feedback being the appropriate {\it class} of ethical
consideration, e.g., ``it is about cheating'', using a small set of categories, and (4) ethics reasoning with user feedback being
natural language.
We find that in all cases, \gptshort's accuracy significantly increases with time, without retraining,
as our approach \csrr{enables it} to use corrective feedback from earlier examples to avoid similar misunderstandings on future examples. In summary, our \textbf{contributions} are:
\reallysquishlist 
\item We show that a large model like \gptshort can be improved after deployment, without retraining, through a memory-assisted architecture. 
\item Our implementation, \ours, is the first demonstration that this is possible - this is an important step forward for real use of LMs, and the paper sets out a general architecture that others can build on, a specific implementation, and detailed evaluation on multiple tasks.
\squishend






\section{Related work}
\label{sec:related}

\emnlpcr{In \citet{interscript}, we show that using a memory of user feedback can be used to repair erroneous model in a supervised setting.}
In this work, we build upon the recent advances in few-shot prompting to modify \gptshort's behavior by adding user feedback to the query (prompt).
Like others, we use \gptshort with {\it few-shot prompting}, where the prompt consists
of a {\bf prefix} $prefix$ containing a few input-output ``training'' examples of the task, followed by the {\bf input} $x$, e.g., a question,
to operate on. However, while prior work has focused on constructing better prefixes, e.g., dynamically selecting good ``training'' examples
based on the question \cite{Scao2021,liu_what_2021}, or even representing the prefix latently \cite{Li2021PrefixTuningOC},
our work elaborates the input $x$ itself to clarify the intended task, by adding user feedback $fb$ from previous misunderstandings. 

\eat{
Our use of recalled memories is a form of ``prompt engineering'', where \gptshort's behavior
is modified by adding to the query (prompt) \cite{Scao2021}. While prior work has added selected QA examples to the prompt (e.g., using KATE \cite{Liu2021WhatMG}), or even 
added continuous vectors \cite{Li2021PrefixTuningOC}, our novel contribution is using a growing repository of user feedback for prompt enhancement.
Further, unlike existing work where the added prompt is fixed after deployment, our prompt can change dynamically at run-time. This further implies that the performance of our model is not fixed, but can instead grow with user interaction.
}

Similarly, our work can be seen as a form of retrieval-augmented QA. Extensive prior work has used retrievals from a text corpus to aid QA, e.g., \citet{Pan2019ImprovingQA,Guu2020REALMRL}, or retrievals of prior QA pairs for nearest-neighbor QA \citep{Khandelwal2020GeneralizationTM}. In contrast, we retrieve from a dynamic memory of user feedback.

The idea of failure-driven reminding and dynamic memory date back several
decades, e.g., \cite{SchankRoger1983DynamicMA,Riesbeck1981FailureDrivenRF}.
Our work resurrects these ideas in a modern context.

Learning from instruction has become important for large LMs that can perform a task based on direct instruction rather
than examples \cite{Wei2021FinetunedLM,Mishra2021NaturalIB}. Our work extends this by adding an adaptive component when those instructions are misinterpreted.
While it may not be possible for a user to provide meaningful feedback on the output itself, giving feedback on the understanding of the instruction is more feasible.

Our approach aims to modify the model's behavior through prompting, given a wrong answer.
An alternative, recently explored approach is ``model editing'' - updating the model
itself by modifying its parameters to fix incorrect answers \citep{mend-mitchell, de-cao-etal-2021-editing, hase2021beleifs}.
Model editing approaches have to date been limited due to uncontrollable out-of-scope changes \cite{mend-mitchell}. In contrast, our goal is not just to correct a prediction, but to generalize that correction
for new problems by collecting feedback to clarify the misunderstanding without damaging the model's basic problem-solving acumen. 

Finally, our work is a simple example of debugging and learning via dialog. While system debugging through dialogue has been explored in many contexts~\citep{Hixon2015LearningKG,Wang2016LearningLG,Davis1977InteractiveTO}, our contribution is a dialogue about the model's understanding of the user's intent.

\section{Approach}
\label{sec:method}


\subsection{Memory enhanced \gptshort architecture}
In our setup, given an input \quesm, a model generates an output \ansm and a sentence \ram expressing its understanding of the task, a skill learned through few-shot examples in the
prompt (Appendix~\ref{sec:actualprompt}). 
The user can then critique \ram by providing natural language feedback \fbm. This is feasible even if the user does not know the correctness of \ansm because they are critiquing the \textit{model's understanding of their intent} rather the answers themselves. 

\begin{table*}[!ht]
    \centering
    \small
    \begin{tabular}{|p{0.19\textwidth}|p{0.43\textwidth}|p{0.3\textwidth}|}
    \hline
    Task (\fbm type) & ($\ques \rightarrow \ans$)  & \ram and \fbm \\ 
    \hline
    Lexical relations  (\instr)   & \quesm: What sounds like good?    &  \ram:  Question is asking for a synonym. \\
                      & \ansm:  wood         & \fbm:  No, I want a homophone. \\  \hline
    Word scrambling (\instr)   & \quesm:  Find the right word given this cycled word: elylarg  &  \ram: The question is about anagram. \\
                      & \ansm:  largely        & \fbm:  No, its about uncycling a word. \\  \hline
    Ethical reasoning (\cat) & \quesm: Turning my blender on at 3AM   &   \ram:  Question is about authority. \\
                      & \ansm: It's bad.         & \fbm:  No, it is about harm. \\ \hline
    Ethical reasoning (\nl)       & \quesm: John has started using again after his mother passed & \ram: Question is about spending money. \\
               & \ansm: It's bad.                                             & \fbm:  No, it is about drug use. \\    \hline
    \end{tabular}
    \caption{Feedback types and demonstration of understanding: our system leverages user feedback to prevent failures caused due to a misunderstanding of the task (\instr) or semantics of the input~(\cat and \nl). We achieve this by having the model articulate an understanding \ram, on which a user can provide feedback using \fbm.}
    \label{tab:tasks-and-fb}
\end{table*}


Given a new query, \ours uses \fbm from similar, prior queries to enrich the (few-shot) prompt \promptm. 
We use the principle that if \csrrcr{two inputs} ${x}_i$ and ${x}_j$ are similar (\ie ${x}_i \sim {x}_j$), then their feedback $\V{fb}_i$ and $\V{fb}_j$ should be exchangeable $(x_i \sim x_j \Leftrightarrow fb_i \sim fb_j)$.
\csrrcr{The underlying assumption here is that for a fixed model, similar inputs will incur similar errors, and thus can use the same feedback for correction.}
Fig. \ref{fig:method} gives an overview of \ours, with the following components:

\paragraph{Memory $\mathcal{M}$}: \memorym is a growing table of key~($\ques_i$) - value~($\V{fb}_i$) pairs that supports read, write, and lookup operations. 
The write operation is used whenever a user gives new feedback.


\vtwo{\paragraph{Lookup \retm}:
The memory allows lookup operations, denoted as \retm, that matches the query=$\ques$ against all the keys of \memorym.}


\vtwo{\paragraph{Combiner $\mathcal{C} (\ques, \memory(\ques))$}: A gating function allowing irrelevant, retrieved feedback to be ignored.}


\paragraph{Few-shot prompting}
Let us briefly recap few-shot prompting with \gptshort. Consider a general setup where given an input \quesm, a model is expected to generate an output \ansm. In a few-shot prompting mode~\citep{Brown2020GPT3}, a prompt \promptm consists of $k$ $(\ques, \ans)$ ``in-context'' examples, i.e., $\prompt = \ques_1 . \ans_1 \sep \ques_2 . \ans_2 \ldots \sep \ques_k . \ans_k$,
where $\sep$ is a token separating examples \csrrcr{and . indicates concatenation}.
During inference, the user inputs a question $\ques_i$, and the model is fed $\prompt\ \sep\ \ques_i$ (\ie the question suffixed to the prompt) and is expected to generate the answer $\ans_i$ as a continuation.

\paragraph{\ours setup} 
\csrrcr{As mentioned, given an input \quesm, we prompt the model to generate an output \ansm and a sentence \ram expressing its understanding of the task.
Thus, the in-context examples for \ours are of the form $\ques \rightarrow \ra, \ans$. 
In addition to the input \quesm, \ours retrieves a \fbm if a question similar to \quesm has been asked before.
To enable the model to react to such feedback, we also include examples of the form \fbsample in the prompt, which are aimed to teach the model to react to $\fb$~(Appendix~\ref{sec:actualprompt}).}

\subsection{Verbalizing Task Understanding}
\emnlpcr{Existing methods for receiving user feedback typically assume the user knows the correct answer \ansm \cite{elgohary-etal-2021-nledit}.
This assumption is paradoxical: if the user knew the answer, why would they be using the model? Further, allowing only ``oracle'' users (who know correct \ansm) might lead to sampling biases. 
In real-world settings, it is common for users to not have the exact answer, but rather, a general understanding of what they are searching for.
Thus, we propose eliciting a verbalization of task understanding \ram from the model in addition to the answer. End users can thus critique \ram. 
}

\emnlpcr{We operationalize this idea by including task verbalization in the prompt (Fig.~\ref{fig:verbalizingexamples}).
Given a question \textit{What sounds like < sighted > ?}, a vanilla prompting approach will generate the answer \textit{cited}.
In contrast, we include a \ram \verbalization{\textit{the homophone for}} in the prompt.
Large-scale language models, such as \gptshort, have been shown to excel at reasoning with a limited number of examples, making them well-suited to mimic the prompt and generate not only the answer, but also an understanding of the task at hand.
Given a test question \textit{What sounds similar to < sighted > ?}, if the model generates \verbalization{\textit{the word that has the same meaning}} as \ram, the user has a reason to believe that the answer is wrong.
Our experiments demonstrate that \gptshort models are able to generate this additional information in all tasks presented.
}

\emnlpcr{Our approach is not foolproof--- the model may spell out a wrong \ram while giving out the correct answer, misleading the user into believing that there is an error (or vice-versa). Hallucinating remains a critical limitation of generative models \cite{cao2022hallucinated}, therefore additional heuristics and model calibration might be necessary to make our approach foolproof. 
In practice, however, we found such cases to be rare for the tasks in this paper.}

\begin{figure}[!ht]
\centerline{
\centering
\fbox{%
    \parbox{0.45\textwidth}{%
	\underline{}

\textbf{(Word reasoning task)\\ Ques:} What sounds like < sighted > ?

\textbf{Ans:} \verbalization{the homophone for} sighted is cited.

\
\hrule
\

\textbf{(Ethical judgment task)\\Situation:} i heard that if you cringe about your past it just means you've grown as a person, for anyone who needs to hear it.

\textbf{Morality Judgment:} \verbalization{This question is about: regretting poor decisions from your past.} The answer is it's okay.

}}}
\caption{\ours is tuned to generate \verbalization{task understanding} + answer. This allows the users to provide feedback on the task understanding even without knowing the actual answer.}
\label{fig:verbalizingexamples}
\end{figure}

\subsection{Allowing GPT-3 to react to feedback}
\emnlpcr{Once the feedback is received from the user, can the model successfully utilize it? By adding a few examples of the form $\ques, \fb \rightarrow \ra, \ans$ in the prompt and setting $\fb=\ra$, we force the model to use the task understanding present in the input when generating the output~(Figure~\ref{fig:reactingtofeedback}). 
Recently, it has been shown that such repetition plays a crucial role in the success of few-shot prompting models~\citep{madaan2022text}.}

\begin{figure}[!ht]
\centerline{
\centering
\fbox{%
    \parbox{0.45\textwidth}{%
	\underline{}

\textbf{Ques:} What is similar to popular ? clarification: when I ask for similar to, I want a synonym.

\textbf{Ans:} \verbalization{the synonym of} popular is admired.

}}}
\caption{An in-context example of the form $\ques, \fb \rightarrow \ra, \ans$, which encourages \ram to be like  \fbm, thereby conditioning the output to react to \fbm. 
}
\label{fig:reactingtofeedback}
\end{figure}

\subsection{Feedback on model's understanding}
\label{sec:feedback}
Within the setup $\ques \rightarrow \ra, \ans$, we focus on following two modes of failure:
\reallysquishlist
\item Task instruction understanding: this is especially concerning in a multi-tasking setup, where the model may consider the question to be about a different task than the one user intended.
\item Task nuanced understanding: when the model understands the task type, but misunderstands the subtle intent in a question. 
\squishend



Our primary goal is to elicit feedback on the model's understanding of the task, however, we also explore settings where an Oracle is available to provide feedback on the labels (as detailed in Section~\secref{sec:webqaexperiments}). 
Finally, we note again that the model reacts to the feedback because some in-context samples are of the form: \fbsample.
We consider a diverse set of tasks ($\ques \rightarrow \ans$), \fbm and \ram, \emnlpcr{as} summarized in Table \ref{tab:tasks-and-fb}.

\subsection{Tasks}
\label{sec:task}
We apply our approach to four tasks: (1) lexical relations (e.g., antonyms, Figure \ref{fig:running-example}),
(2) word scrambling (e.g., anagrams), (3) ethics (with user feedback being the appropriate {\it class} of ethical
consideration), and (4) ethics (with user feedback being natural language). 
For all five tasks, the dataset consists of \fbsample tuples, where \fbm clarifies the task in \quesm.
We have a simulated conversational setting, in which a user can ask the model \quesm (covering any of these five tasks). If the model gives a wrong answer to query \quesm, then \fbm is used as the simulated corrective feedback.
The sources for these datasets are listed in Appendix ~\secref{sec:source}.

\subsubsection{Lexical Relations}

The lexical relation task is to predict a word with a given lexical relationship to an input word.
We use five relationships: synonym (\textit{syn}), antonym (\textit{ant}), homophone~(\textit{hom}), definition (\textit{defn}), and sentence usage generation (\textit{sent}).

\subsubsection{Word Scrambling}
For this task, given a word with its characters transformed, the model is expected to recover the original characters.
There are four transformation operations the user can request:  reversal of words (\textit{rev}, yppup $\rightarrow$ puppy), cycle letters in word (\textit{cyc}, atc $\rightarrow$ cat), random insertions (\textit{rand}, c!r ic/ke!t$\rightarrow$ cricket), and anagrams by changing all but the first and last (\textit{anag1}, eelhpnat $\rightarrow$ elephant) or all but the first and last 2 characters (\textit{anag2}, elapehnt $\rightarrow$ elephant).
We use the original dataset by \citet{Brown2020GPT3}.\footnote{word scrambling dataset \url{https://github.com/openai/gpt-3/tree/master/data}}

For both these tasks, each question can be asked in multiple ways~(\eg for synonym generation, the users might ask questions of the form \textit{what is like}, \textit{what has a similar sense}, \textit{what is akin to}, \textit{what is something like}, etc.)
Similarly for the lexical relations task, we specify the task description $x$ using different phrasings, e.g., ``rearrange the letters'' (which the system sometimes misunderstands), and the (simulated) user feedback $fb$ is a clearer task description, e.g., ``The anagram is''. The system thus accumulates a set of ($x$, $fb$) pairs in memory after each failure, helping it avoid future misunderstandings of $x$ through feedback retrieval.



\subsubsection{Ethical Reasoning (2 tasks)}

For ethical reasoning, we consider a setup where given a situation~(\eg \textit{cheating on your partner}), the model is expected to provide a judgment on whether the situation is ethical or not~(\eg \textit{it's not okay}).
In addition to providing a judgment on the ethics of the situation, the model also elucidates its understanding of what the question is about~(\eg \textit{being loyal}).
While the user may not know the answer, we posit that they would be able to provide feedback on the broader context.
For example, if the model generates \textit{being financially savvy} instead of \textit{being loyal} for the situation \textit{cheating on your partner}, a user can still point out this problem and provide feedback.


We use a subset \footnote{social norms dataset (social-chemistry-101, \citet{forbes2020social})  \url{https://github.com/mbforbes/social-chemistry-101}} of the dataset provided by~\delphi~\citep{jiang2021delphi}. We simulate two different kinds of user feedback, using two of the
annotations attached to each example in the Delphi dataset:

\reallysquishlist
\item Categorical feedback~(\ertcat): In this setting, the model generates its understanding $u$ of the situation by selecting one of 10 different possible categories of morality to which the situation might belong: \textit{care, loyalty, authority, fairness, sanctity, degradation, cheating, subversion, betrayal, and harm}.
These categories are explicitly provided for each example in the Delphi dataset.

\item Natural language feedback~(\ertnl): For this, we use the associated ``rule of thumb'' (RoT) annotation —a  general moral principle — attached to each example in the Delphi dataset. 
To compile a challenging subset of the data for \ertnl, we sample by input length, preferring long \quesm, with a short feedback \fbm. 
Specifically, we use the top 1\% of the inputs by length to create a challenging set of input situations~(\quesm).
\csrr{User feedback \fbm is a natural language feedback on the understanding \ram.}
\squishend

\csrr{In both the cases, the model is ``taught'' to generate a category \ram (as well as the okay/not-okay answer \ansm to the ethical question) by being given a few examples in the prompt prefix, thus articulating which moral category (for \ertcat) or rule-of-thumb~(for \ertnl) it thinks is applicable. The simulated feedback \fbm is the gold category associated with the example in the question, if \gptshort gets the answer wrong.}

We selected these tasks because situations that involve reasoning about similar ethical principles can utilize similar past feedback. For example, \textit{sharing an extra umbrella with your friend if they don't have one}, and \textit{donating surplus food to the homeless} both involve \textit{compassion}.


\begin{figure}[t]
\centering
    \includegraphics[scale=0.25]{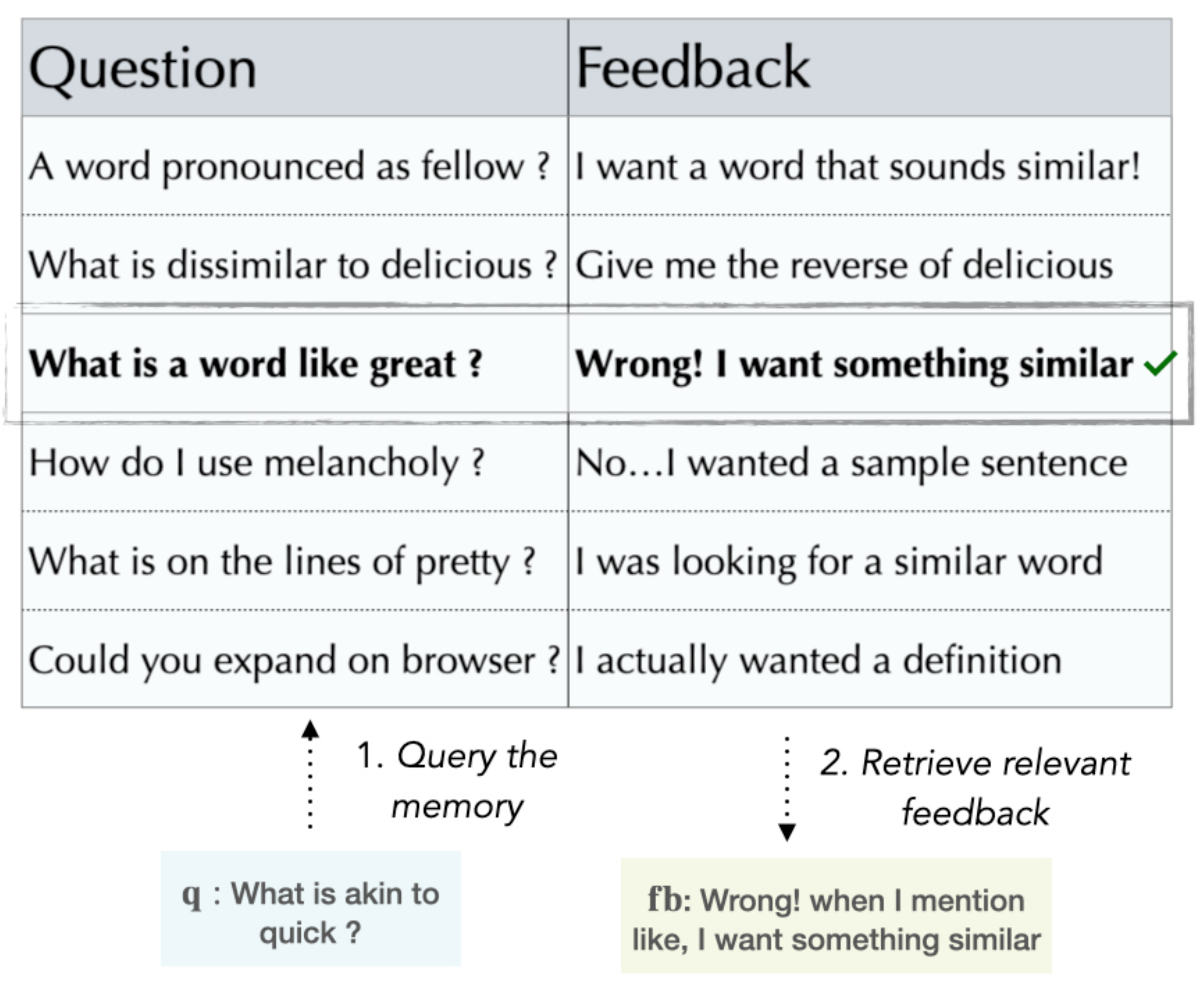}
    \caption{Sample snapshot of memory for lexical QA.}
    \label{fig:memsample}
\end{figure}

\subsection{\ours Implementation}

\paragraph{Implementation of memory \memorym } 
\memorym uses the user input \quesm as the key and the corresponding feedback \fbm as value.
Given a question $\ques_i$, if the user detects that the model has misunderstood the question, they may provide a $\fb_i$ with \textit{clarification probability} \fprobi.
The ($\ques_i$, $\fb_i$) pair is stored in a memory \memorym, with $\ques_i$ as the key and $\fb_i$ as the value.
For a subsequent question $\ques_j$, the retriever \retm checks if a similar question appears in memory. If yes, then the corresponding feedback is attached with the question and fed to the model for generation.

For example, a question asking for a synonym, such as \textit{what is akin to fast?} might be misinterpreted as a request for antonyms.
As mentioned, in our setup, the model generates its understanding of the task \ram, and not just the answer to the question.
The user, by inspecting \ram = \textit{The opposite of fast is:} might determine that the model has misunderstood them, and give feedback \textit{i wanted a synonym}, which gets stored in \memorym.
If a similar question~(\eg \textit{what is akin to pretty ?}) is asked later by the same or a different user, the corresponding feedback~(\textit{i wanted a synonym}) is attached with the question to generate the answer. Figure \ref{fig:memsample} illustrates a sample memory for this task.

\paragraph{Implementation of retriever \retm} 

\vtwo{A retrieved past feedback that is incorrect might cause the model to make a mistake, thus necessitating a good retrieval function. We propose a two-stage method for effective retrieval involving: transforming \quesm, followed by a similarity lookup of the transformed \quesm in \memorym. When the task involves high surface-level similarity among past feedback, such as in lexical word tasks, then a simple heuristic-based transformation is sufficient.
However, such simple transformations are insufficient for tasks that involves more complex retrieval e.g., when two lexically dissimilar situations can share the same understanding. 
For example, consider two situations from \ertnl: \textit{Filling a false time sheet at work} and \textit{Being at a party, and telling parents I am studying}. 
These situations look lexically dissimilar but correspond to the same underlying social principle \textit{lying to authority.}
In our experiments, off-the-shelf methods failed to address these challenges~(see \secref{sec:experiments} later). 

To address these challenges with transformation in complex tasks, we have designed a novel \sts based transformation called \ourir. Given \quesm, \ourir generates a \textit{transformed} feedback $\hat{\fb}$ for \quesm using a \textit{generative} \sts model. Our approach is inspired and supported by the recent success of generate and retrieve \cite{mao2021generation} methods.
However, despite the similarity, the methods have different goals: \citet{mao2021generation} leverage generative models for query expansion, whereas our goal is explainable input understanding. 
See Appendix~\ref{sec:generativeir} for more details on \ourir.

After the transformation stage, the closest matching entry is then used as the corresponding \fbm. Transformation reduces  $\memory(\ques)$ to a search over $\fb_1, \fb_2, \ldots, \fb_{|\memory|}$ with $\hat{\fb}$ as the search query. We compute similarity based on a fine-tuned Sentence transformers~\citep{reimers-2019-sentence-bert}. 
}

\paragraph{Implementation of combiner $\mathcal{C}$} $\mathcal{C}$ concatenates \quesm with relevant \fbm retrieved by \retm.  \vtwo{To ensure that the \quesm is appended with \fbm only if it is relevant, our current implementation of combiner uses a threshold on the similarity score between the \quesm and the closest feedback \fbm retrieved by \retm.}
\vtwo{We rely on the model (\gptshort) to pay attention to the relevant parts of the input. Exploring more complex gating mechanisms remains an important future work.}



\section{Experiments}
\label{sec:experiments}


\paragraph{Baselines} 
We compare \ours (memory-assisted prompt editing) with two baselines:
\reallysquishlist
    \item \textbf{\nomem} This is the standard \gptshort\footnote{We use \gpt~(davinci) for all experiments.} in few-shot prompting mode~(hyper-parameters listed in {Appendix~\secref{sec:hyperparams}}). Input is $\prompt\ \sep\ \ques_i$ (\ie question $\ques_i$ appended to prompt $\prompt$).
    It generates answer $\ans_i$ and its understanding of the user's intent $\ra_i$.
    \item \noindent\textbf{\growprompt:} Similar to $\nomem$, but the $\prompt$ is continuously grown with a subset of memory $\memory$ that can fit within the prompt (max. 2048 tokens).
    The most recent subset of $\memory$ of memory inserted is inserted in the prompt.
    The ethical reasoning tasks~(\ert) involve long examples, and the initial prompt itself takes close to the max allowed tokens. 
    Thus, the \growprompt setup is only provided for the lexical relations and word scrambling tasks.
\squishend

\paragraph{Metrics}


We use two different metrics:

\reallysquishlist
\item $Acc(\ans)$: \% of cases where answer matched the ground truth.
\item $Acc(\ra)$:  \% of cases where the model's understanding of user's intent is correct. $Acc(\ra)$ is also referred to as instruction accuracy.
As discussed in ~\secref{sec:feedback}, depending on the task, the model generates its understanding on either the instruction or semantics of the question.
\squishend

\paragraph{Clarification probability} 
In real-world cases, we cannot expect a user to provide feedback for all the examples (\eg the user might not know that the understanding of the model is wrong).
To simulate this realistic setting, we experiment with various values of clarification probabilities $Pr$.


\subsection{\ours improves \gptshort accuracy} 
Does pairing \gptshort with \ours help? \csrr{\secref{subsec:results_ethical_tasks} empirically validates this on ethical reasoning tasks and  \secref{subsec:results_word_tasks} on word reasoning tasks.}

\subsubsection{Ethical reasoning tasks}
\label{subsec:results_ethical_tasks}

Table \ref{tab:resultsert} presents results on the \delphi dataset (1,000 points in the test set). Recall from \secref{sec:task} that there are two kinds of feedback on \delphi questions: \cat and \nl feedback. \ours gets over 25\% relative improvement for both \ertnl and \ertcat.
\csrrcr{We found that having an efficient retriever was critical for \ertnl: sentence transformer based retriever scored 38.5, vs. 45.2 using \ourir, a 17\% improvement.}

\begin{table}[!h]
    \centering
    \small
\addtolength{\tabcolsep}{-3pt}    

\begin{tabular}{lrr}\\ \toprule
model       & \ertcat     & \ertnl    \\ \hline
\nomem      &  48.3 &  34.4       \\
\ours       &  \textbf{60.0} & \textbf{45.2}  \\ \bottomrule
\end{tabular}%
\addtolength{\tabcolsep}{3pt}

\caption{\ours outperforms \nomem for both the categorical and the more challenging \ertnl setup having longer, ambiguous inputs.}
\label{tab:resultsert}
\end{table}

\begin{figure}[!h]
    \centering
    \includegraphics[width=\columnwidth]{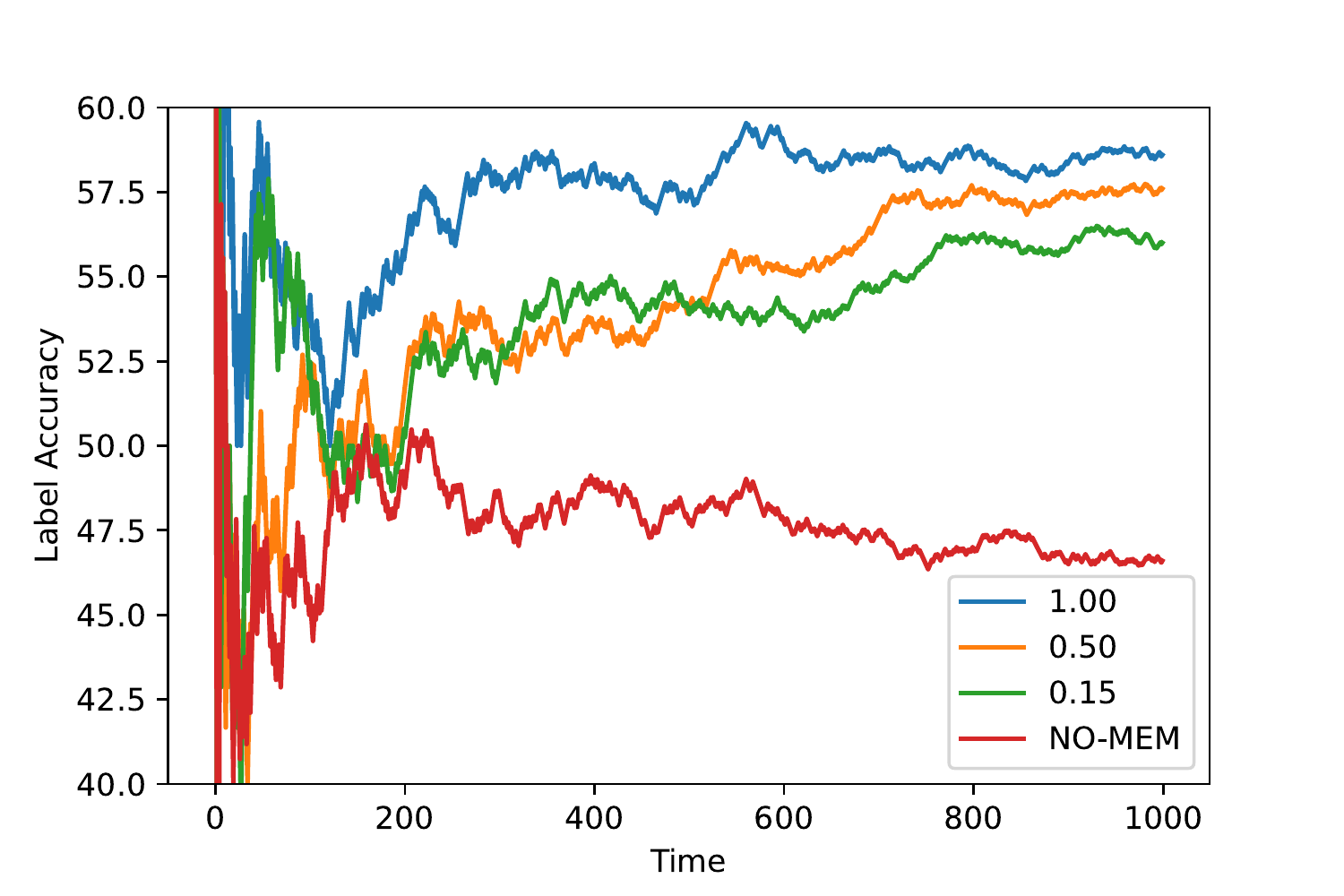}
    \caption{\ertcat: Label accuracy increases with time for all values of clarification probabilities \fprobi.}
    \label{fig:delphicataccuracy}
\end{figure}

\begin{figure}[!h]
    \centering
    \includegraphics[width=\columnwidth]{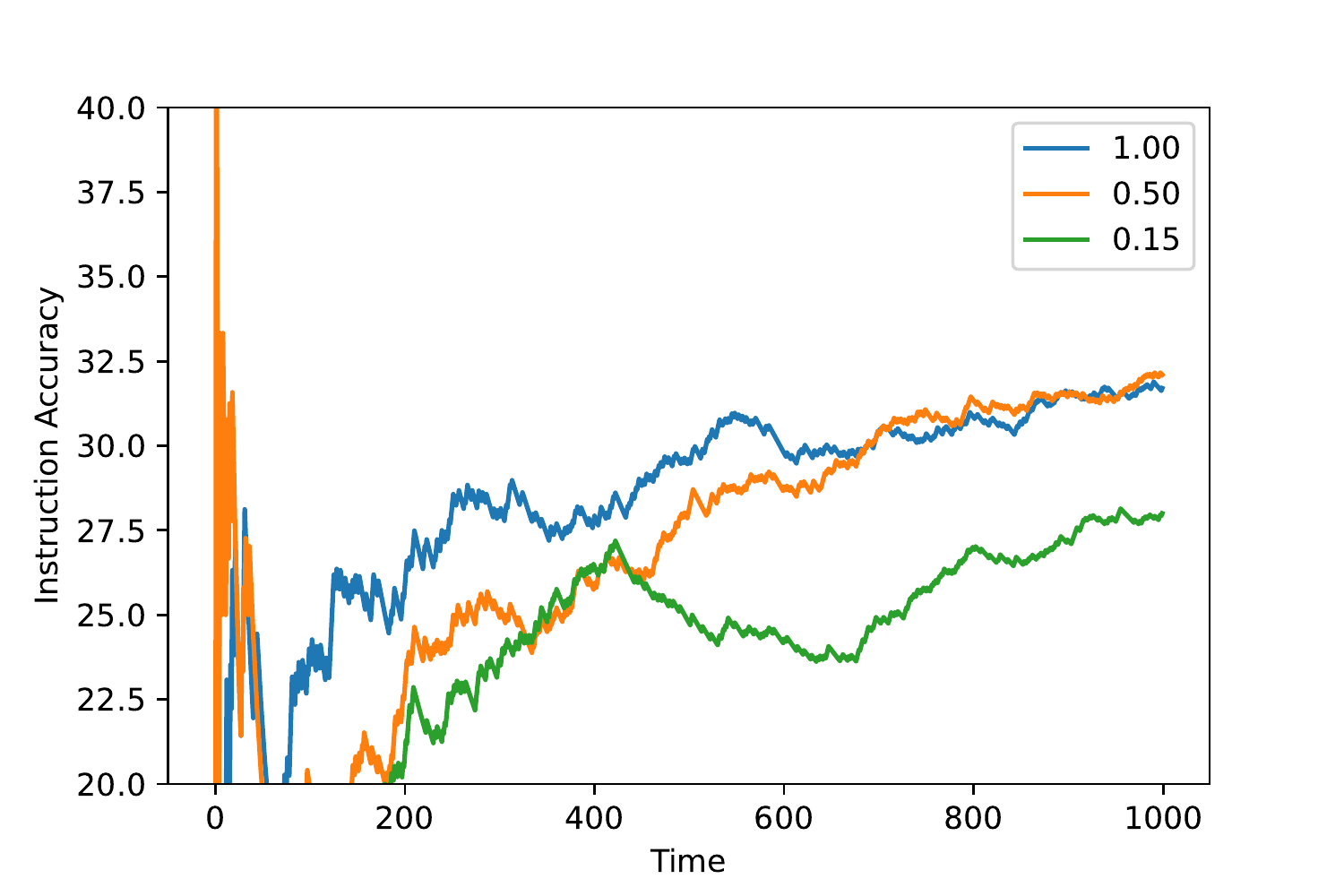}
    \caption{\ertcat: \csrrcr{Instruction accuracy sharply increases with a larger clarification probability, showing that \ours responds to feedback. With time, lower values of \fprobi catch up as memory is gradually filled with error cases and feedback.}}
    \label{fig:delphiinstraccuracy}
\end{figure}





\begin{table*}[!t]
    \centering
    \small
    \begin{tabular}{|p{0.12\textwidth}| p{0.36\textwidth} | p{0.2\textwidth} | p{0.23\textwidth} |}
    \toprule
    Error category  & Input   & Output (\ours)   & Reference \\
    \midrule
    Correct, lexically variant understanding     & when I said "i do" to my wife, I am now realizing I only meant "i'll try". does she deserve to know and a chance to back out? & This question is about: telling a spouse your true feelings.  & This question is about: loving your partner.\\
    \midrule
    Distracted understanding & i am 20 years old, lately I have been forgetting simple spellings while taking notes in class . I often bump into random objects around hurting myself. & This question is about: wanting to kill yourself. The answer is \bad. & This question is about: seeking medical attention if you feel there is something wrong with you. The answer is \good. \\     
    \midrule
    Retrieval failures & checked out my high school ex girlfriends instagram for the first time in two years after graduating. I'm happy I did. she’s not as pretty as she used to be and I feel great.
        & [retrieved feedback] This question is about: looking at someone's social media. The answer is \good.
        & This question is about: judging people based on physical attributes. The answer is \bad.  \\
    \hline
    \end{tabular}
    \caption{\ert \nl task error categories}
    \label{tab:error-analysis-ert-nl}
\end{table*}

\paragraph{\csrr{\ours effectively incorporates feedback, improving accuracy over time}}
Figure \ref{fig:delphiinstraccuracy} demonstrates that the instruction accuracy increases over time \csrrcr{for different values of clarification probability.}

Fig. \ref{fig:delphicataccuracy} shows that label accuracy improves over time. Baseline (\nomem) saturates after 200 time steps; \ours continues to improve. 
Continuous improvement is one of our key advantages.
These charts show that instruction accuracy and label accuracy are correlated~\csrr{(corr. coeff = 0.36)}.

\csrrcr{We observe that using a higher clarification probability leads to a sharp increase in instruction and label accuracy early on in the training for both \ertcat and \ertnl. This is because a higher clarification probability causes the feedback memory to fill up more quickly, providing more feedback for new questions.} 

\paragraph{Error analysis: Ethical-\nl} In \ert \nl and \cat tasks, a primary source of label errors is confusion between labels such as \okay and \good due to the nuanced differences e.g., input = teaching your child a musical instrument. \ours predicts \good, but the expected answer is \okay. \citet{jiang2021delphi} make similar observations.


We randomly sampled examples from the \ertnl dev set where the model generates an incorrect understanding~(i.e., $Acc(\ra)=0$ based on exact match).
Our goal is to understand the typical errors made by the model and use the analysis to calibrate the findings in Table~\ref{tab:resultsert}.
We select \ertnl for the analysis because it involves free-form natural language which is difficult to study quantitatively.


\reallysquishlist
\item \textbf{Correct, lexically variant understanding (30\%)}: 
Exact match underestimates model performance (as the task involves generation). $\sim$ 30\% \ram is a lexical variation of the reference gold understanding. E.g., \textit{telling a spouse your true feeling} vs. \textit{loving your partner}. The generated label in these 30\% cases is still correct. 
(Table~\ref{tab:error-analysis-ert-nl}, row 1)
\item \textbf{Distracted understanding (50\%)}: A major source of instruction and label errors is the model getting distracted by an unimportant context. 
Bad retrieval accounts for 30\% errors within this category, \eg matching a situation in the memory where the expected understanding is only partially applicable to the query. (Table~\ref{tab:error-analysis-ert-nl}, row 2)
\item \textbf{Retrieval failures (18\%)}: These errors are caused by an irrelevant retrieved understanding from the memory \vtwo{, when using a state-of-the-art retrieval method (Table~\ref{tab:error-analysis-ert-nl}, row 3). 
\ourir helps to reduce these retrieval failures. 
See Appendix~\secref{sec:generativeir}}. 
 
\squishend
Table \ref{tab:error-analysis-ert-nl} presents canonical examples of these error categories. We also find that over time, more relevant past examples are fetched (see Table \ref{tab:neighbors-ert-cat}).



\subsubsection{Word Reasoning Tasks}
\label{subsec:results_word_tasks}

For these tasks, we compare gold $\ra^*$ and generated \ram based on hard-coded linguistic variations (\eg \textit{the antonym is} matches \textit{the opposite is}).
While we do not explicitly evaluate task accuracy, we observe a near-perfect correlation between the accuracy of \ansm and \ram~(\ie if the \gptshort understands the task correctly, the output was almost always correct).
\csrrcr{This shows improving model's understanding of a task might lead to an improved performance.}

\begin{figure}[!b]
    \centering
    \includegraphics[width=\columnwidth]{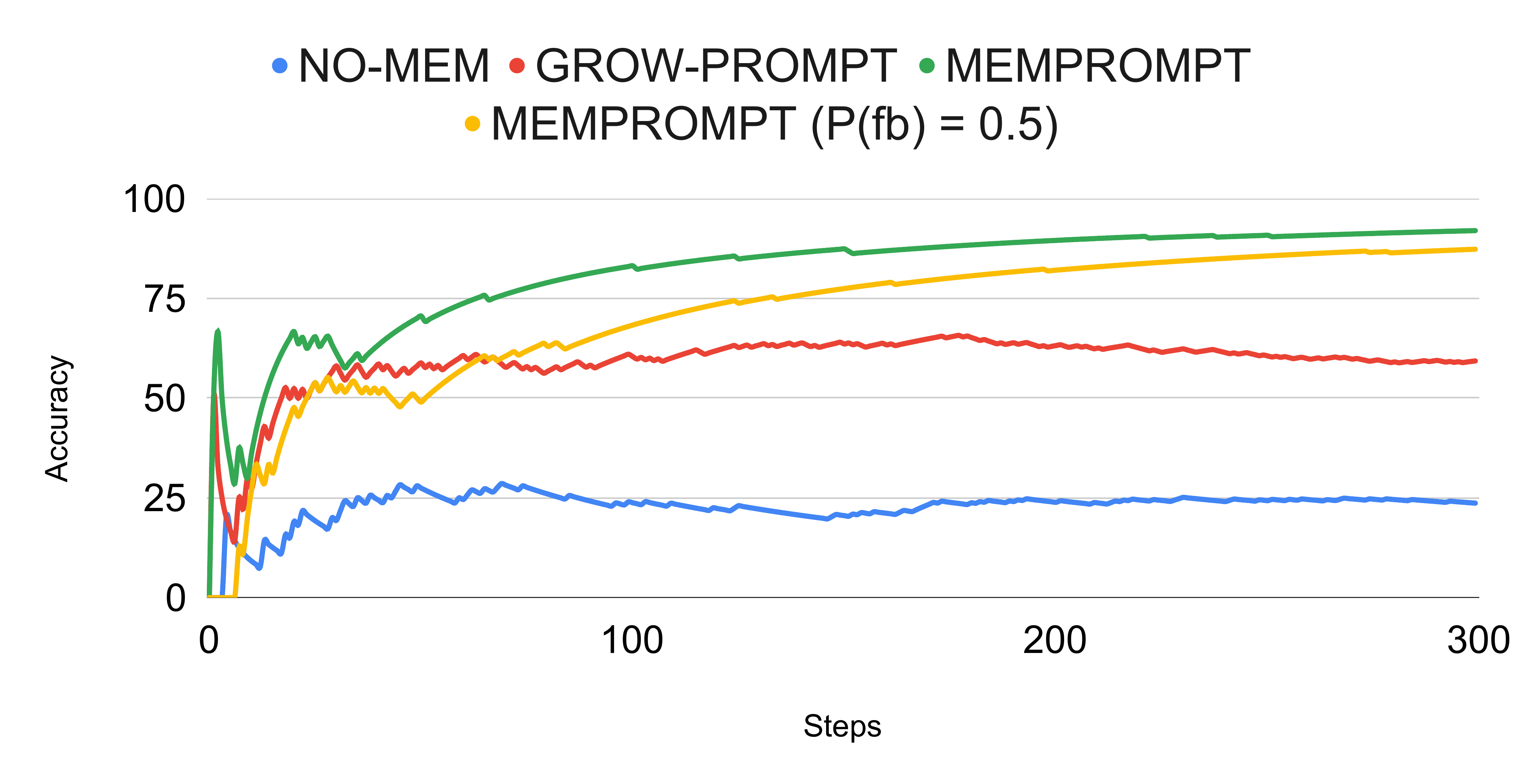}
    \includegraphics[width=\columnwidth]{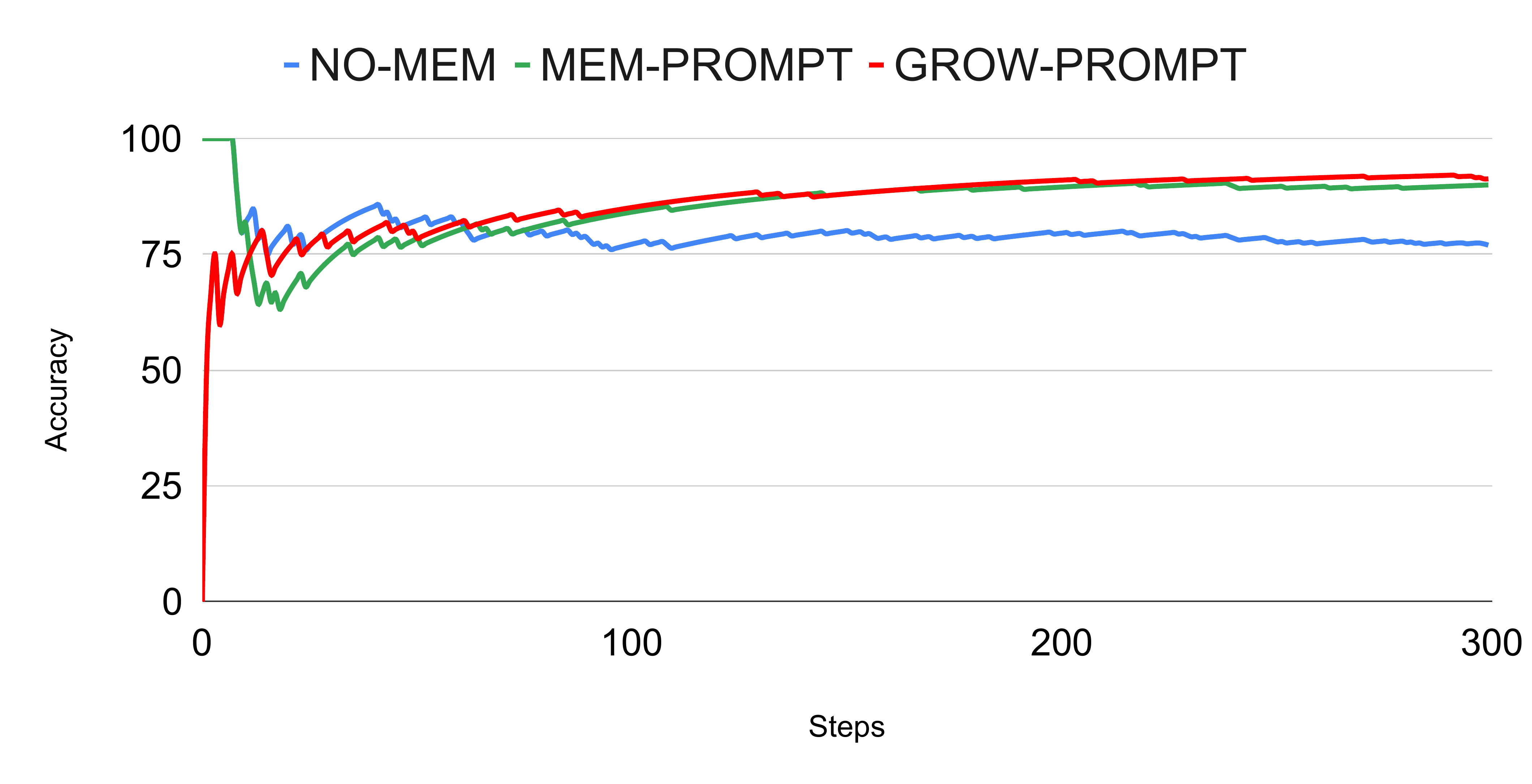}
    \caption{Avg. performance on lexical (top) and word scramble (bottom) tasks with time (x-axis).
    Accuracy increases with time as memory is filled up with feedback from past errors.}
    \label{fig:main-result}
\end{figure}

Figure \ref{fig:main-result} reports the overall performance on the word reasoning tasks. 
The accuracy improves substantially within 300 examples when using memory (in yellow) vs. no memory (in blue). 
Note that our approach operates in a few-shot learning regime, where there is no pre-existing training data available. The only examples provided to the model are through the prompt.
The performance of \growprompt (red) lies in between, showing that non-selective memory is partially helpful, although not as effective as failure-driven retrieval (our model).
However, \growprompt is $\sim$ 3x more expensive~(larger prompts) and cannot scale beyond the 2048 tokens limit. 
We also found that the retrieved feedback from memory was effective 97\% of the time; only in $\approx$ 3\% of cases feedback had no positive effect. 



When the memory is used for every example (green line, Fig \ref{fig:main-result}, top), the performance improves quickly vs. the yellow line~(\fprobi = 0.5).

\begin{table}[!ht]
    \centering
    \small
    \addtolength{\tabcolsep}{-3pt}    

    \begin{tabular}{lrrrrrr} \\ \toprule
model  &    syn  &  ant   &  hom   &  sent  &  defn  &  all \\ \hline
\nomem  &  0.58  &  0.43  &  0.13  &  0.30  &  0.39  &  0.37 \\
\growprompt  &  0.71  &  0.87  &  0.75  &  0.92  &  0.76  &  0.80 \\
\ours  &  \textbf{0.99}  &  \textbf{0.98}  &  \textbf{0.98}  &  \textbf{0.98}  &  \textbf{0.96}  &  \textbf{0.98} \\ \bottomrule
\end{tabular}
\addtolength{\tabcolsep}{3pt}
    \caption{Results on lexical qa: \ours has the best performance across all lexical \qa tasks.}
    \label{tab:results}
\end{table}

\begin{table}[]
    \centering
    \small
\addtolength{\tabcolsep}{-3pt}    
\begin{tabular}{lrrrrrr}\\ \toprule
model       & anag1         & anag2         & cyc           & rand       & rev           & all           \\ \hline
\nomem      & 0.81          & 0.47          & 0.95          & 0.98          & 0.62          & 0.77          \\
\growprompt & \textbf{0.86} & \textbf{0.89} & 0.93          & \textbf{0.96} & 0.90          & \textbf{0.91} \\
\ours       & 0.81          & 0.83          & \textbf{0.98} & 0.95          & \textbf{0.93} & 0.90        \\ \bottomrule  
\end{tabular}%
\addtolength{\tabcolsep}{3pt}

\caption{\growprompt and \ours outperform \nomem on all word scramble \qa tasks.}
\label{tab:resultsword}
\end{table}


\subsection{Using dynamic prefix in prompts} 
\csrr{Recent work such as \citet{liu_what_2021} investigate using dynamic prompts for better generation. For a given input \quesm, their method(~\kate) relies on retrieving examples from the training set that are similar to \quesm for dynamically creating the prompt \promptm. Note that our method edits \quesm with a feedback \fbm, and is thus complementary to \kate.
To demonstrate this, we conduct experiments on \ertcat and \ertnl tasks, where dynamic prompts were created using \kate, and \ours was used to attach feedback to the question. Our results show a consistent 10\% improvement when using both \kate and \ours, indicating that the improvements are complementary.}

\subsection{\ours with label feedback}
\label{sec:webqaexperiments}

\ours requires the model to verbalize its understanding of the question, on which a user provides feedback. 
To investigate the efficacy of \ours in settings where generating an understanding is not easy, we experiment with factual question answering on the \webqa dataset~\citep{berant2013semantic}, and find that \ours is effective even with label feedback (Appendix~\secref{sec:webqaexperimentsappendix}).






\subsection{\csrr{Using \ours for language and dialects based personalization}}
\csrr{We demonstrate an application of \ours for personalization with a use-case where user language preferences can be folded in the memory. We simulate a user who does not speak fluent English and uses code-mixed language. The queries posed by the user contain words from two Indian languages: Hindi and Punjabi. \gptshort predictably misunderstands the task. The user clarifies the meanings of their dialect/language phrases. While initial queries fail, subsequent queries that reuse similar words succeed because their clarifications are present in the memory (details in Appendix~\secref{sec:lowresourceappendix}).}





\section{Conclusion}

\eat{We design a simple, and novel memory-enhanced \gptshort that allows users to interact and improve the model without retraining. This work opens the door to a new generation of machines that can be dynamically taught by interacting with people, rather than statically finding patterns in pre-provided datasets, potentially allowing millions of users to personally instruct and refine their AI agents. 
}
We present \ours, a novel, memory-enhanced \gptshort that allows users to interact and improve the model without retraining. A key insight is to have the model articulate not just its answer but also its understanding of the user's intent, providing an avenue for feedback. 
We show that deployed systems with fixed large-language models can still be improved by interacting with end-users, potentially improving their performance and broadening their utility.

\section*{Acknowledgments}
We thank Dheeraj Rajagopal and Yannic Kilcher for the insightful and engaging discussions.
This material is partly based on research sponsored in part by the Air Force Research Laboratory~(agreement number FA8750-19-2-0200). 
The U.S. Govt. is authorized to reproduce and distribute reprints for Governmental purposes notwithstanding any copyright notation thereon. 
The views and conclusions contained herein are those of the authors and should not be interpreted as necessarily representing the official policies or endorsements, either expressed or implied, of the Air Force Research Laboratory or the U.S. Government.

\section{Limitations}

We have shown how to improve very large models through interaction. Our memory-based enhancement is a low-cost utility enhancement eventually geared towards personalized, correctable models, which is currently an open question in NLP with unresolved issues. While our method is a step toward a promising open direction, it comes with limitations and opportunities when deploying to the real world.

\paragraph{Scaling} In practical deployments of the \ours method, the memory can grow to orders of magnitude, introducing scaling challenges. We anticipate using memory as a buffer between cycles of re-training, and these cycles could range from a week to several months. Between cycles of re-training, \ours can serve as a way to avoid repeating mistakes and collect feedback which can be used to fine-tune and improve the next version of the model.

Currently, we operate with \textit{a single user} at a time, but a real-world deployment could encounter multiple users. These users could exhibit characteristics of a user community where some feedback could apply to multiple users in a community cluster, while others differ in interpretation and style. In such a multi-user environment, managing the memory effectively when dealing with incompatible entries would be important. Existing initial ideas towards managing a bank of beliefs could be extended to address these problems, e.g., \cite{kassner2021beliefbank}. In addition, when looking up such a rich and potentially noisy feedback collection, rather than retrieving a single feedback item, it would help to have an adapter over the memory that generates feedback by adapting the existing, diverse, and related past feedback to the current scenario. This increases the diversity of the generated knowledge and reduces the impact of erroneous feedback and noise. 

\paragraph{Ethical concerns}
Extending the discussion on noise in feedback, our setting assumes that users will not provide any \textit{adversarial} feedback. However, in real-world environments, this assumption is unlikely to hold. Additionally, there is a risk in the real-world deployment of our system, wherein an adversarial user might provide harmful feedback, thus maliciously controlling the systems (potentially a home-based robot) where our method is deployed. Thus, robust mechanisms such as \ourir and memory adapters will be critical for successful real-world deployments.

Privacy is another ethical concern, as the deployed system collects and records feedback from a user, some of which could contain personal information (\textit{when I look for an interesting movie, I mean something that contains romance}). Therefore, the system needs to win the trust of the users so they would be encouraged to interact closely, and to win this trust, the system needs to demonstrate smartness, receptivity to user feedback, and the ability to maintain the memory without leaking any personal information safely. 

Finally, large-language models generate text that might be biased and insensitive to a user's socio-cultural context~\citep{bordia2019identifying,sharma2021evaluating,hovy2021five}.
In a multi-user deployment of our system, the memory could contain feedback from user communities of diverse beliefs, gender identities, and cultural backgrounds could lead to conflicts. Thus the system will need checks and balances to ensure that the content produced by the system as a result of the feedback is not harmful.

\bibliographystyle{acl_natbib}
\bibliography{custom}
\newpage
\clearpage

\appendix

\section{Inside \ours: Populating and using the memory}

\begin{figure*}[]
    \centering
    \includegraphics[scale=0.245]{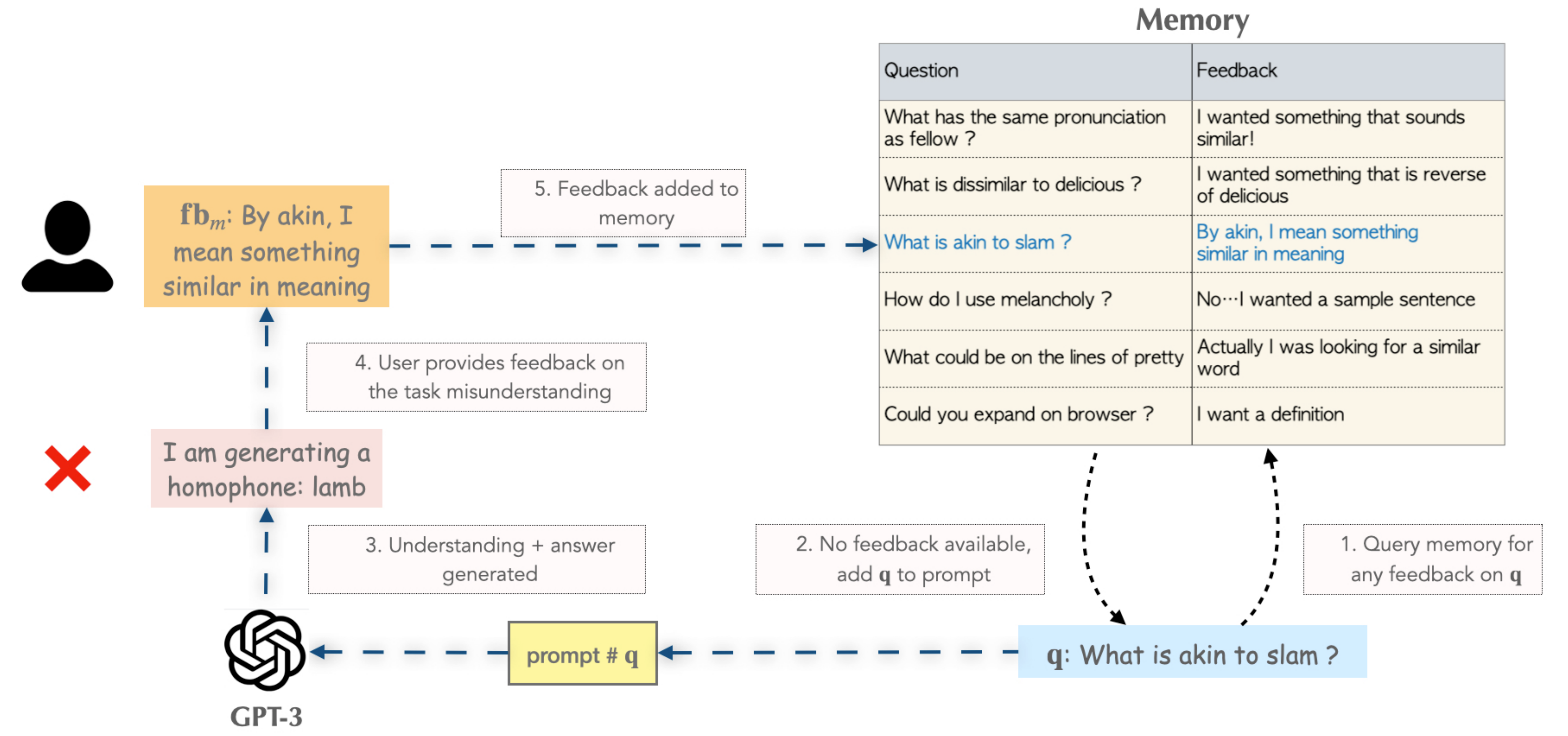}
    \caption{\ours: adding to memory. User enters a question for which no feedback is available (steps 1, 2). Directly prompting \gptshort with the question leads to incorrect answer and understanding (step 3). User-provides feedback on the incorrect understanding (step 4), which is added to memory (step 5).}
    \label{fig:flow-populate}
\end{figure*}

\ours maintains a growing memory of recorded cases where a feedback was provided to clarify the user's misunderstood intent. This flow is presented in Figure \ref{fig:flow-populate} that shows a sequence of steps 1-5 on how the memory is populated.

\begin{figure*}[!t]
    \centering
    \includegraphics[scale=0.26]{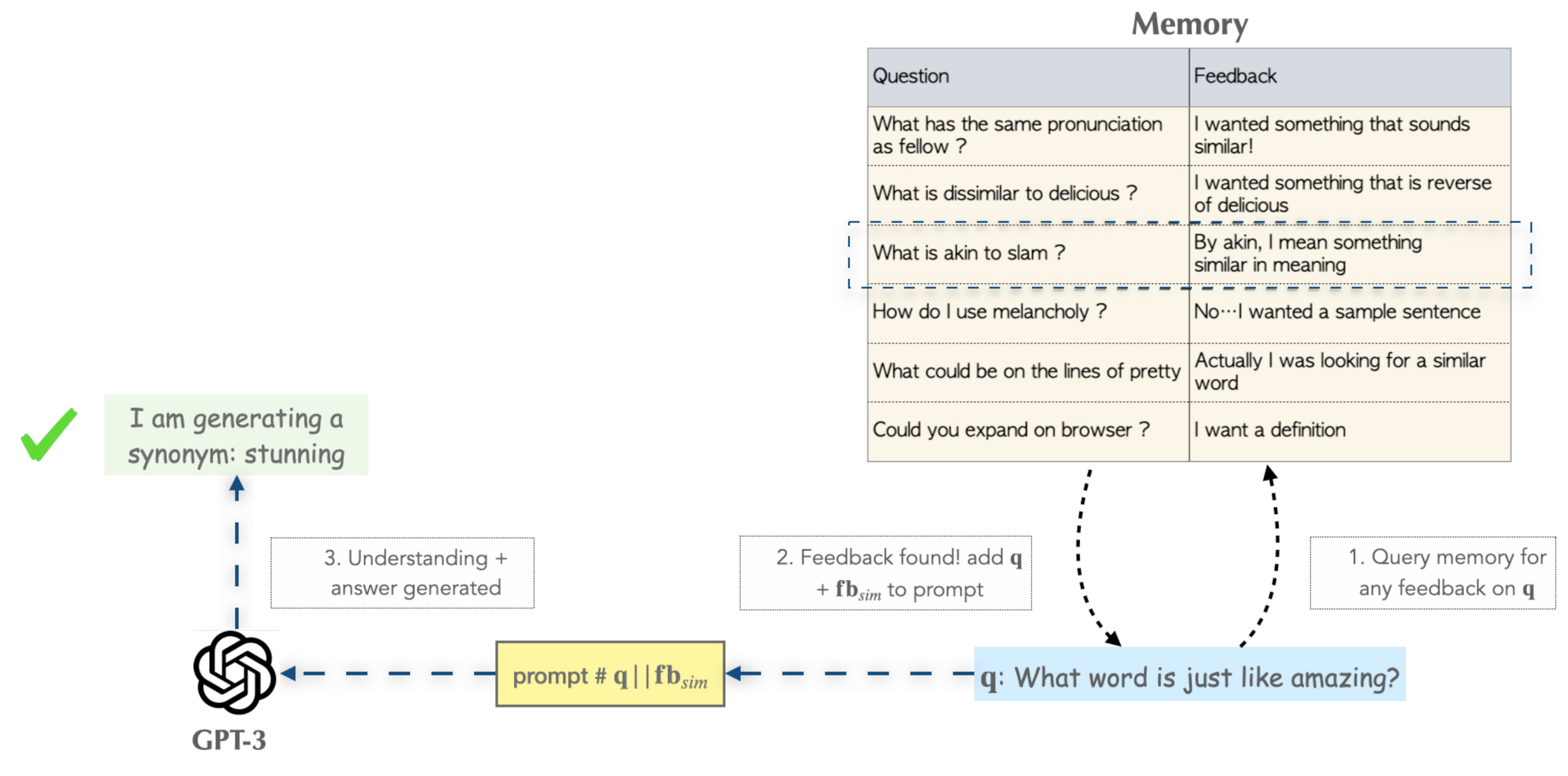}
    \caption{\ours: retrieving feedback from memory. User enters a question which \gptshort has incorrectly answered in the past, and has received feedback from a user (step 1). The feedback is retrieved from memory (step 2), and both question and feedback are added to the prompt. The prompt contains examples that allow~\gptshort to react to user feedback and generate correct understanding and answer.}
    \label{fig:flow-retrieve}
\end{figure*}

\ours also produces enhanced prompts for any new query based on the user feedback on similar cases recorded previously in the memory. Figure \ref{fig:flow-retrieve} presents the sequence of steps 1-3 involved in retrieving and applying a past feedback on a similar case.

\section{Generative IR (\ourir)}
\label{sec:generativeir}
\textit{A note on feedback and understanding} Feedback \fbm and understanding \ram are two concepts that we repeatedly use in this work.
Briefly, \ours requires a model to spell out its understanding of the instruction (\ram).
The user can then provide a feedback \fbm on the understanding.
In the prompt, both \fbm and \ram are identical.
Such examples are of the form $\ques, \ra \rightarrow \ra, \ans$ and their main purpose is to reinforce that model the input feedback \ram be used to generate the output.

\subsection{Introduction}

\newcommand{\si}{$s_i$\xspace}
\newcommand{\sj}{$s_j$\xspace}
\newcommand{\ci}{$\fb_i$\xspace}
\newcommand{\cj}{$\fb_j$\xspace}
\newcommand{\va}{\mathbf{a}}
\newcommand{\vb}{\mathbf{b}}

One of the key strengths of \ours is its ability to leverage feedback provided on earlier inputs \quesm to improve a current input.
This is achieved by retrieving a feedback from memory \memorym using \quesm as the key.
An underlying assumption of this process is that similar inputs will admit similar feedback, allowing us to use the feedback provided for one situation on another.
For two input situations \si and \sj with respective feedback \ci and \cj, this assumption can be succinctly stated as:
$$s_i \sim s_j \implies \fb_i \sim \fb_j $$

The ethical reasoning dataset with natural language feedback, \ertnl, \emnlpcr{provides a unique challenge for this assumption because lexically dissimilar situations might have the same feedback.} 
As a concrete example, consider an input situation
\si : \textit{tom hated skating because he had no sense of balance} -- with a feedback \ci : \textit{this question is about practicing more when you want to improve your skills}.
Suppose that our system has already seen \si and has received a feedback \ci (\ie there is an entry in \memorym: \si $\rightarrow$ \ci).
Next, suppose a user enters a new situation \sj: \textit{jordyn was trying to improve her soccer skills}.
As usual, \ours will try to retrieve feedback for a \textit{similar} situation.
However, such retrieval is going to be challenging, because \si~(\textit{tom hated skating because he had no sense of balance}) has little to no overlap with \sj~(\textit{jordyn was trying to improve her soccer skills}), although humans can easily tell that both situations are about improving skills.
Consequently, \ours may fail to retrieve the relevant feedback \ci or worse, may retrieve a misleading feedback. 

\emnlpcr{The fact that two ostensibly dissimilar inputs two inputs $(\ques_i, \ques_j)$ may share the same feedback makes vanilla retrieval non-viable for our setting.
We deal with this challenging situation with two different solutions of increasing complexity.}

\subsection{Initial approach: Learning a feedback similarity function}
\label{sec:irfinetune}
Since the surface level similarity of input situations is not enough to capture similarity of respective feedback, we attempt to learn a function $f_{\theta}$ that will map similar inputs $\ques_i$ and $\ques_j$ to similar representations if the corresponding feedback $\fb_i$ and $\fb_j$ are close to each other, and vice-versa.
A natural choice is training an embedding function $f: \ques \rightarrow \mathrm{R}^d$ supervised by $\texttt{cos}(\fb_i, \fb_j)$ where $\texttt{cos}$ is the cosine similarity~($\texttt{cos}(\va, \vb) = \frac{\va^T\vb}{|\va||\vb|}$).
Thus, the objective function is:
$$
    \mathcal{L}_{\theta} = (\texttt{cos}(f_{\theta}(\ques_i), f_{\theta}(\ques_j)) - \texttt{cos}(\fb_i, \fb_j))^2
$$

Intuitively, this objective function will encourage the similarity between the inputs~($\texttt{cos}(f_{\theta}(\ques_i), f_{\theta}(\ques_j))$) to be high when the corresponding feedback are similar, and vice-versa.

Feedback retrieval proceeds as follows: an input \si is embedded using $f_{\theta}$, and $f_{\theta}(s_i)$ is then used to retrieve a feedback from the memory, with the hope that representations $f_{\theta}(s_i)$ and $f_{\theta}(s_j)$ will be similar after the training.

While in principle this objective function should be enough to learn informative representations that bring two inputs with similar feedback close, we found the training to be unstable.
We attribute this to the fact that two extremely dissimilar situations can have identical feedback.
\emnlpcr{Given limited training data, it might be unrealistic to train similarity functions that can capture all possible cases where the same feedback applies to two situations.}
As a way to circumvent this, we also experiment with a generative version of our method, described next.

\begin{figure*}[!ht]
    \centering
    \includegraphics[width=\textwidth]{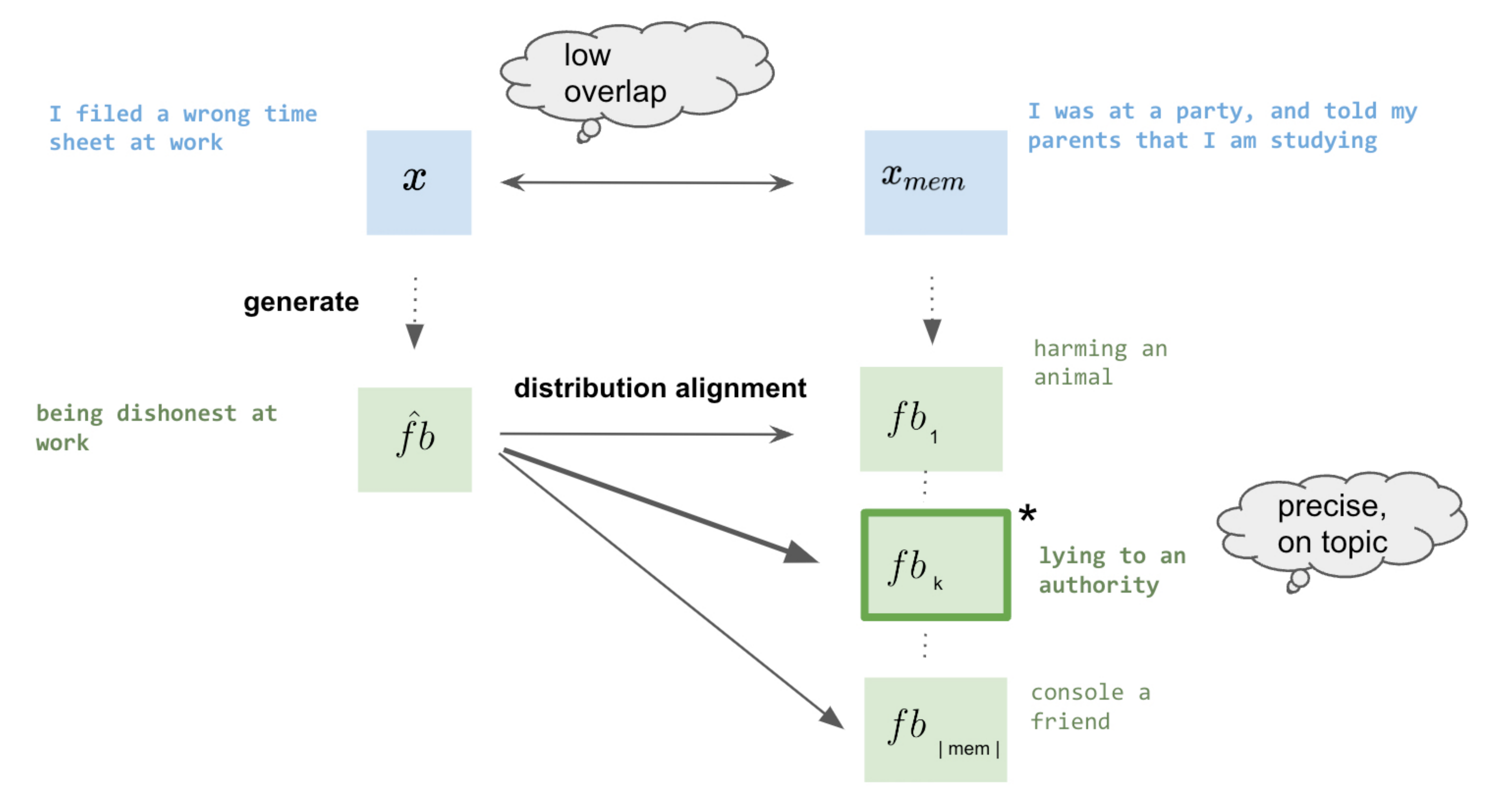}
    \caption{\textbf{Overview of \ourir}. To retrieve a relevant feedback that applies to \quesm, \ourir first generates a feedback $\hat{\fb}$ using a generative model. This is then aligned with a corpus of feedbacks $\fb_1, \fb_2, \ldots, \fb_{|tr|}$ (\eg sourced from the train split). The best matching feedback $\hat{\fb^*}$ is then used for \quesm.
    Thus, \ourir decomposes the retrieval problem $\ques \rightarrow \fb$ into two sub-problems: (i) generate a rough feedback ($\ques \rightarrow \hat{\fb}$) and (ii) search for the closest feedback in a large store
    $\hat{\fb^*} = \argmin_{j \in [1, |tr|]} |\hat{\fb} - \fb_j|$.
    }
    \label{fig:ir-overview}
\end{figure*}

\subsection{Proposed approach: Training generative model for retrieving similar feedback}

To address these retrieval issues, we propose \textsc{\textbf{GUD-IR}} (Generated UnDerstanding for explainable IR). 
The key intuition for our approach relies on substituting $f_{\theta}: \ques \rightarrow \mathrm{R}^d$ (latent space projection) with $f_{\theta}: \ques \rightarrow \fb$ (generated understanding of \quesm). 
Concretely, instead of learning a function that maps a question to a $d$ dimensional vector, we train a generative model that directly maps an input to a rough understanding.
The generated rough understanding is then used as a key to retrieve a relevant understanding from the database using any off-the-shelf retrieval method. 
This two-step \textit{generate-then-retrieve} procedure has benefits: (i) it alleviates sparsity issues that we found latent space projection methods were unable to deal with\footnote{\eg there are only eight popular emotions but can lead to a large number of diverse situations. Hence, many inputs can map to the same principle \fbm. This mapping becomes increasingly difficult for a model as the specificity of \fbm increases, because of sparsity issues. This is exacerbated when the input situations are diverse and previously unseen.} (ii) the overall retrieval becomes explainable and debuggable.  

Our approach is inspired and supported by the recent success of generate and retrieve \cite{mao2021generation} methods.
However, despite the similarity, the methods have different goals: \citet{mao2021generation} leverage generative models for query expansion, whereas our goal is explainable input understanding. 
Moreover, their implementation is geared towards open-domain QA, while ours is towards explainable input understanding. Thus, it is non-trivial to adapt similar ideas to our tasks effectively.




Specifically, we train a \sts model, (\eg \tf~\citep{raffel2020exploring}), that maps each input \quesm to a corresponding output \fbm.
The feedback is now retrieved in a two step process:
\squishlist
    \item [1.] The generative model $f_{\theta}$ is used to generate a noisy feedback for \si, $\hat{\fb}$.
    \item [2.] $\hat{\fb}$ is used as a key to \textit{search} over the set of already present feedbacks, to retrieve the nearest one.
\squishend
Instead of directly using clarification to lookup the nearest feedback, we first transform the input to the space of clarifications, then search over the set of already present clarifications. 
Figure \ref{fig:ir-overview} presents an overview of our \textit{generation then reshape} approach (\ourir).
As we discuss in Section~\ref{subsec:results_ethical_tasks}, \ourir was key to achieving good performance for the \ertnl task.

\begin{figure*}[!t]
\minipage{0.33\textwidth}
  \includegraphics[width=\linewidth]{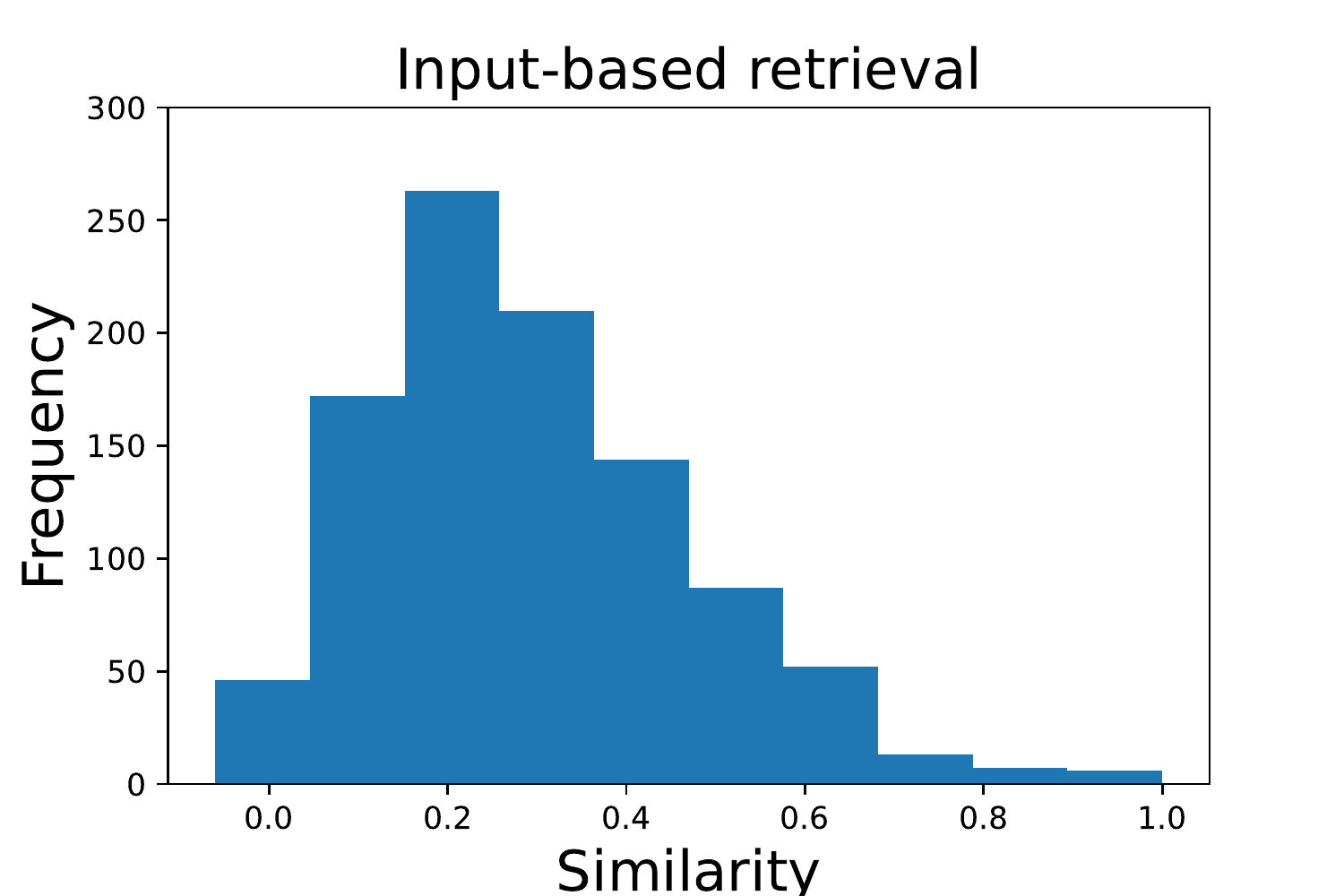}
\endminipage\hfill
\minipage{0.33\textwidth}
  \includegraphics[width=\linewidth]{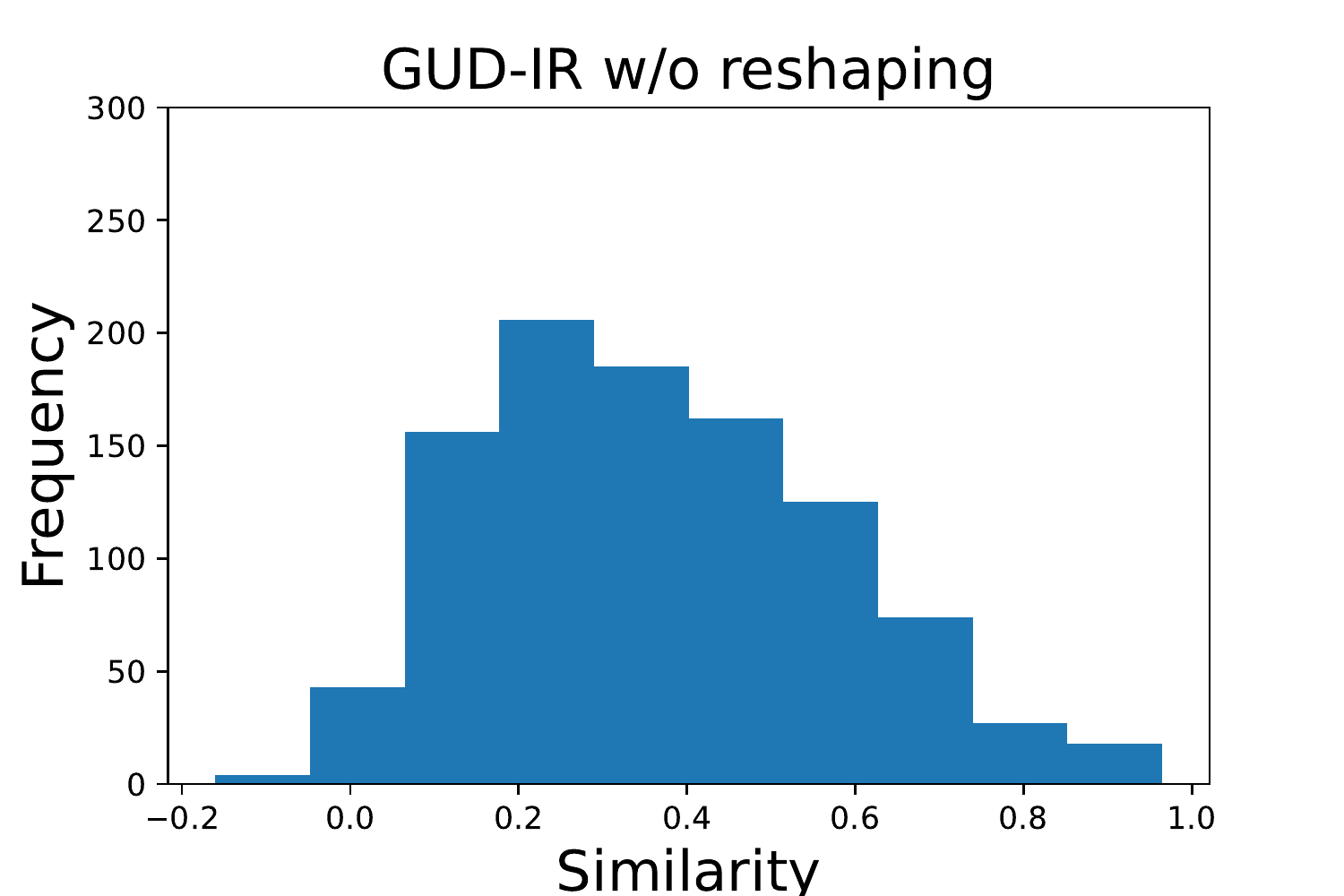}
\endminipage \hfill
\minipage{0.33\textwidth}
  \includegraphics[width=\linewidth]{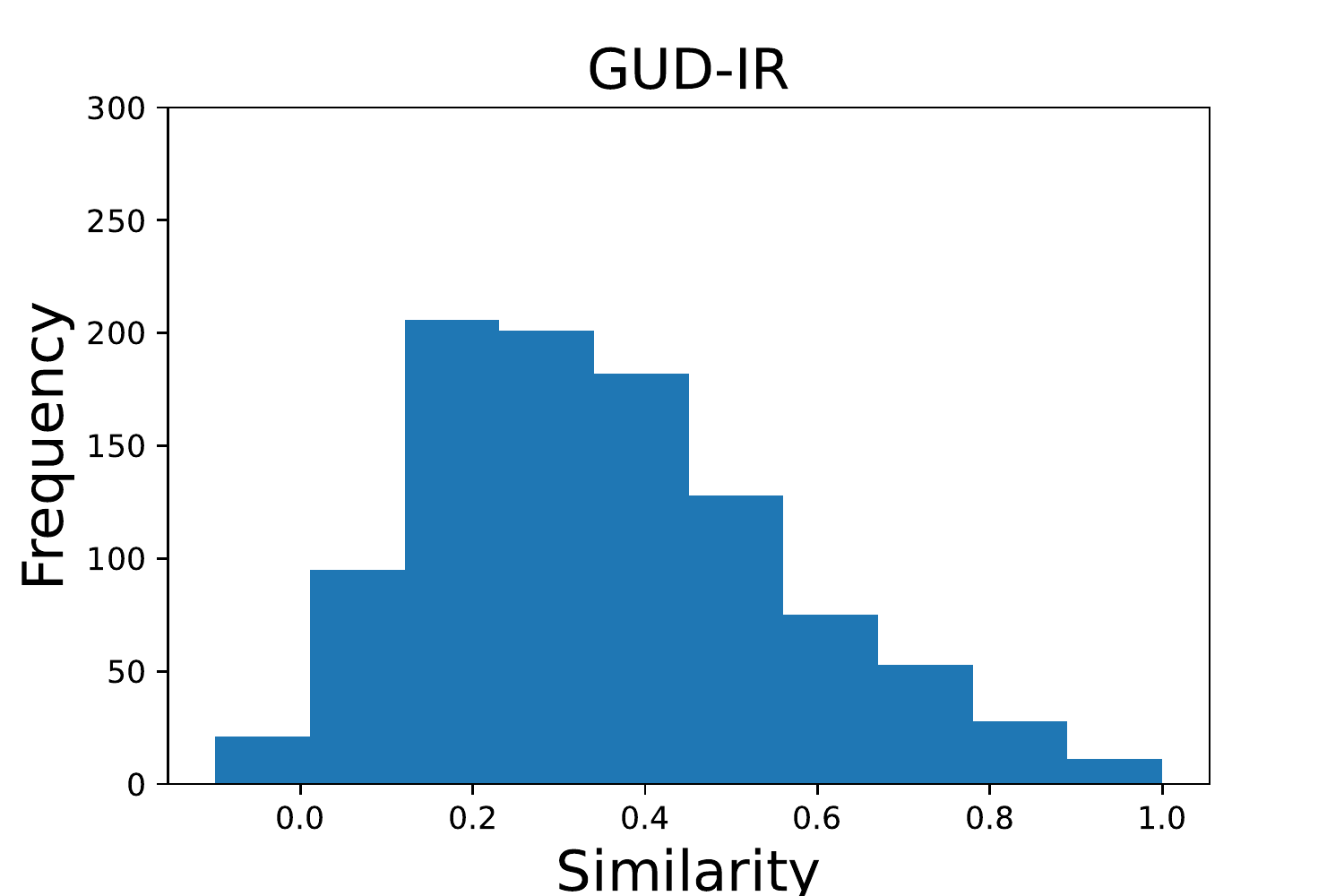}  
\endminipage \hfill
\caption{Distribution of similarity scores between expected \fbm$^*$ and $\hat{u}$ for retrieval (left) and \ourir (right). The similarity scores are higher using \ourir. 
}
\label{fig:simdistirbution}
\end{figure*}

In addition to the task accuracy, we plot the distribution of $\texttt{sim}(\hat{u}, \hat{u}^*)$~(similarity of the true and retreived feedback) over the test set for different retrieval methods.
Figure~\ref{fig:simdistirbution} shows this distribution using \ourir and using surface-level similarities. The probability mass shifts towards a higher similarity range for \ourir.

\csrrcr{The lexical reasoning and \webqa tasks present a simpler setting for retrieval, as similarity of keys indicates a similarity of values. For such cases, we use Sentence transformers~\citep{reimers-2019-sentence-bert} to encode the query, and cosine similarity with a threshold of 0.9 to find a matching key.}

\section{Querying \gpt using OpenAI API}
\label{sec:hyperparams}
We use the OpenAI API for querying \gpt.\footnote{\url{https://beta.openai.com/docs/introduction}, we use `text-davinci-001`}
The python code is listed below.
Here, ``PROMPT'' is set to prompt shown in~\secref{sec:actualprompt}, followed by the input question \quesm and feedback \fbm if applicable.

We used a temperature of 0.0 for factual \qa (\webqa) experiments to select the most likely token at each step, and this setting does not require generating diverse answers, as one would expect for a factual domain. For \ertcat and \ertnl, we found that a higher temperature ($\sim$ 0.7) was causing a large divergence in the performance (a difference of $\pm 10$\% accuracy across runs), making reproducibility challenging -- similar observations were made by \cite{summers2021can}. Thus, we used to a temperature of 0.0 for \ert experiments. A temperature of 0.7 was used for all the other experiments.

{\small
\begin{verbatim}
import os
import openai

openai.api_key = os.getenv("OPENAI_API_KEY")

response = openai.Completion.create(
  engine="davinci",
  prompt="PROMPT",
  temperature=0.7,
  max_tokens=64,
  top_p=1,
  frequency_penalty=0,
  presence_penalty=0
)
\end{verbatim}
}

\newpage
\clearpage

\section{Prompt}
\label{sec:actualprompt}

GPT3 is queried using a prompt $\prompt$ of example i/o behaviors,
followed by the actual question \quesm and (optionally) retrieved feedback \fbm.
It then generates the understood intent \ram and answer \ansm as a continuation.
\ram and \ansm are expressed a single sentence, e.g., 
"[The synonym for <word> is] [<word>]"
Figure~\ref{fig:actualprompt} shows this prompt $\prompt$, containing a mixture of
$(\ques \rightarrow \ra, \ans)$ and $(\ques, \fb \rightarrow \ra, \ans)$ "training" tuples.


\begin{figure*}[!ht]
\centerline{
\centering
\fbox{%
    \parbox{0.9\textwidth}{%
	\underline{}
\\
\vspace{2mm}
What is the homophone for < wring > ?

\#

 the homophone for wring is ring END

\#

how do you use < highway > in a sentence?

\#

 a sentence with highway is: soldiers stand guard along the [ highway ] END

\#

can you define < camisole > ?

\#

 the definition of camisole is a sleeveless undergarment. END

\#

What is the antonym for < prohibition > ?

\#

 the antonym for prohibition is permit END

\#

What is the synonym for < surrogate > ?

\#

 the synonym for surrogate is substitute END

\#

how do i use < fog > ?  | clarification: when i ask for how do i use, i want a sentence.

\#

 a sentence with fog is: a rising sun burns the [ fog ] off a city END

\#

What sounds like < sighted > ? | clarification: when I ask for sounds like, I want a homophone.

\#

 the homophone for sighted is cited END

\#

what is like < provident > ? | clarification: when I ask for like, I want a synonym.

\#

 the synonym for provident is prudent END

\#

can you define < rider > ? | clarification: when i ask for define, i want a definition.

\#

 the definition of rider is a person who is riding something. END

\#

What is the opposite of < citation > ? | clarification: when I ask for opposite, I want an antonym.

\#

 the antonym for citation is award END
    }%
}}
\caption{The prompt used for our tasks. 
During inference, an input question $\ques_i$, and optionally a feedback $\fb_i$ is appended after this prompt, and the model is expected to generate the answer $\ans_i$ and its understanding of the question intent $\ra_i$ as a continuation.
The prompt contains examples of the form $(\ques \rightarrow \ra, \ans)$,
expressed "\quesm \# \ram \ansm END \#", 
and $(\ques, \fb \rightarrow \ra, \ans)$, 
expressed "\quesm | clarification: \fbm \# \ram \ansm END \#".
(\ram and \ansm are expressed together as a single sentence, e.g., "[The synonym for <word> is] [<word>].")}
\label{fig:actualprompt}
\end{figure*}

\begin{figure*}[!ht]
\centerline{
\centering
\fbox{%
    \parbox{0.8\textwidth}{%
	\underline{}
\\
\vspace{2mm}
Find the right word after removing random letters from < t!r/e/a/s/u/r.e!s >

\#

 the word after removing symbols from t!r/e/a/s/u/r.e!s is treasures END

\#

Find the original word after ignoring the punctuation and spaces in < e >

\#

 the word after removing symbols from e is elders END

\#

Find the right word given this cycled word: < lprovisiona > ?

\#

 the uncycled version of lprovisiona is provisional END

\#

Make a word while keeping the first and last char < vosiin > ?

\#

 the anagram 1 for vosiin is vision END

\#

Find the original word that is interspersed in < f.i.n!e/p.i/x >

\#

 the word after removing symbols from f.i.n!e/p.i/x is finepix END

\#

Find the right word given this rotated word: < cturalarchite > ?

\#

 the uncycled version of cturalarchite is architectural END

\#

Find the original word after ignoring the punctuation and spaces in < s.e!n.t.i.n/e/l >

\#

 the word after removing symbols from s is sentinel END

\#

Find the right word given this rotated word: < ibitioninh > ?

\#

 the uncycled version of ibitioninh is inhibition END

\#

Figure out the word which has the same first two and the last two char < watsed > ? | clarification: when I want you to figure out the word which has the same first two and the last two char, I mean anagram 2.

\#

 the anagram 2 for watsed is wasted END

\#

Make a word while keeping the first and last char < isucnase > ? | clarification: when I want you to make a word while keeping the first and last char, I mean anagram 1.

\#

 the anagram 1 for isucnase is issuance END

\#

Unscramble everything except the first two and the last two char < acotrs > ? | clarification: when I want you to unscramble everything except the first two and the last two char, I mean anagram 2.

\#

 the anagram 2 for acotrs is actors END

    }%
}}
\caption{The prompt used for our word scrambling tasks derived from \citet{Brown2020GPT3}}
\label{fig:actualpromptwordscramble}
\end{figure*}

\begin{figure*}[!ht]
\centerline{
\centering
\fbox{%
    \parbox{0.9\textwidth}{%
	\underline{}
\\
\textbf{Situation:} last day of high school food fight and I nailed a special needs girl square in the face with a heavy muffin.

 \textbf{Morality Judgment:} This question is about: getting into food fights with people. The answer is It's bad. END

\#

\textbf{Situation:} when phone shopping at stores, I used to write inappropriate stuff and post it to random peoples book of faces accounts who would forget to logoff on the display phones. This question is about: messing with people's social media accounts.

 \textbf{Morality Judgment:} This question is about: messing with people's social media accounts. The answer is It's bad. END

\#

\textbf{Situation:} i had an uncomfortable talk with my 10yo about puberty and feel like avoiding him for a few days.
 
 \textbf{Morality Judgment:} This question is about: being honest with your kids. The answer is It's good. END

\#

\textbf{Situation:} every time I see my three year old nephew being funny or I'm having a good time with him, my mind goes to a school shooting or car crash or kidnapping or some other horrible atrocity happening to him. I have absolutely no idea how to handle this and I'm begging for help.
 
 \textbf{Morality Judgment:} This question is about: admitting you imagine bad things happening to a child and that you need help with this. The answer is It's good. END

\#

\textbf{Situation:} i see someone who’s not there. he’s in my head. he screams sometimes when I'm too stressed and the only way I can get him to leave me alone, is when I cut myself .
 
 \textbf{Morality Judgment:} This question is about: asking someone close to you who is having mental health issues how you can help them. The answer is It's okay. END

\#

\textbf{Situation:} i heard once that if you cringe about your past it just means you've grown as a person, for anyone who needs to hear it. This question is about: regretting poor decisions from your past.
 
 \textbf{Morality Judgment:} This question is about: regretting poor decisions from your past. The answer is It's okay. END

\#

\textbf{Situation:} my cat hopped up in my lap, cleaned herself and went to sleep. now I'm a captive and I want to move but I can't. This question is about: waking up an animal because you would like to move.
 
 \textbf{Morality Judgment:} This question is about: waking up an animal because you would like to move. The answer is It's okay. END

\#

\textbf{Situation:} if it wasn't for the pain it would cause to my family, I would have thrown myself off a bridge a while ago.
 
 \textbf{Morality Judgment:} This question is about: caring about your family's feelings. The answer is It's good. END

\#

    }%
}}
\caption{The prompt used for \ertnl task. The user provides a situation and the model generates a morality judgement. In addition to the morality judgment, the model also provides a high-level understanding of what the task is about.}
\label{fig:ertnl}
\end{figure*}

\begin{figure*}[!ht]
\centerline{
\centering
\fbox{%
    \parbox{0.9\textwidth}{%
	\underline{}
\\

 \textbf{Situation}: i am selectively racist towards "generic" indian recruiters because they waste my time with absolute zero results.

 \textbf{Morality Judgment}: This question is about: trating you differently because of your race. The answer is It's bad. END

\#

 \textbf{Situation}: i told my husband I didn't feel well so I could stay in bed and play red dead redemption 2 all day and night. This question is about: playing hooky every once in awhile.

 \textbf{Morality Judgment}: This question is about: playing hooky every once in awhile. The answer is It's bad. END

\#

 \textbf{Situation}: I've misunderstood the phrase ‘when the going gets tough the tough get going’ for my entire life. This question is about: not understanding cliches when young.

 \textbf{Morality Judgment}: This question is about: not understanding cliches when young. The answer is It's okay. END

\#

 \textbf{Situation}: when phone shopping at stores, I used to write inappropriate stuff and post it to random peoples book of faces accounts who would forget to logoff on the display phones. This question is about: messing with people's social media accounts.

 \textbf{Morality Judgment}: This question is about: messing with people's social media accounts. The answer is It's bad. END

\#

 \textbf{Situation}: i had an uncomfortable talk with my 10yo about puberty and feel like avoiding him for a few days.

 \textbf{Morality Judgment}: This question is about: being honest with your kids. The answer is It's good. END

\#

 \textbf{Situation}: every time I see my three year old nephew being funny or I'm having a good time with him, my mind goes to a school shooting or car crash or kidnapping or some other horrible atrocity happening to him. I have absolutely no idea how to handle this and I'm begging for help.

 \textbf{Morality Judgment}: This question is about: admitting you imagine bad things happening to a child and that you need help with this. The answer is It's good. END

\#

 \textbf{Situation}: i heard once that if you cringe about your past it just means you've grown as a person, for anyone who needs to hear ito. This question is about: regretting poor decisions from your past.

 \textbf{Morality Judgment}: This question is about: regretting poor decisions from your past. The answer is It's okay. END

\#

 \textbf{Situation}: my cat hopped up in my lap, cleaned herself and went to sleep. now I'm a captive and I want to move but I can't. This question is about: waking up an animal because you would like to move.

 \textbf{Morality Judgment}: This question is about: waking up an animal because you would like to move. The answer is It's okay. END

\#

 \textbf{Situation}: if it wasn't for the pain it would cause to my family, I would have thrown myself off a bridge a while ago.

 \textbf{Morality Judgment}: This question is about: caring about your family's feelings. The answer is It's good. END
    }%
}}
\caption{The prompt used for \ertcat task. The user provides a situation and the model generates a morality judgement. In addition to the morality judgment, the model also provides a high-level understanding of what the task is about.}
\label{fig:ertcat}
\end{figure*}



\newpage
\clearpage

\section{Datasets for lexical question-answering tasks}
\label{sec:source}
As mentioned in Section~\secref{sec:experiments}, we focus on five different linguistic $\qa$ tasks.
The source of data for each of these tasks is listed below:
\begin{enumerate}
    \item The synonyms (\syn) and antonyms~(\ant) were obtained from~\citet{nguyen2016integrating}.\footnote{\url{https://www.ims.uni-stuttgart.de/en/research/resources/experiment-data/lexical-contrast-dataset/}}
    \item The homophones~(\homn) were obtained using homz~\url{https://github.com/cameronehrlich/homz}. We use the closest homophone returned by homz for each word in the English dictionary.
    \item The definitions~(\defn) were sourced from \textit{The Online Plain Text English Dictionary}~\url{https://github.com/eddydn/DictionaryDatabase}
    \item Examples for usage in a sentence~(\sent) are from Commongen~\cite{lin2020commongen}.
\end{enumerate}

\subsection{Templates}
We manually created 15 task templates with three variants of phrasing the question for each task. Sample templates are shown in code listing \ref{code1}.
The data (word1, word2) in the code is initialized with the entries in the four sources mentioned above.
The complete file is available in the project repository~\url{https://github.com/madaan/memprompt/tree/main/src/templates}.

\subsection{Sample questions}
Tables~\ref{tab:linguistictasks}, \ref{tab:hinditasks}, and \ref{tab:punjabitasks} list some sample \quesm-\ansm for settings where the question was asked as a linguistic variation, in Hindi, and in Punjabi, respectively. 

\section{\ours with label feedback}
\label{sec:webqaexperimentsappendix}

Our current approach requires the model to verbalize its understanding of the question, on which a user provides feedback.
Such a setup might not be possible, for instance, due to the nature of questions.
Can \ours be effectively used in such settings as well?
To investigate this, we experiment with factual question answering on the \webqa dataset~\citep{berant2013semantic}, and use the test set provided by~\citet{berant2013semantic} for all experiments~(2032 questions).
The \webqa dataset consists of factual questions~(\textit{which language is spoken in Canada?}) with multiple answers~(\textit{English, French}), and is a popular dataset for benchmarking the performance of \gptshort on question answering in a few-context prompting setup.

\begin{table*}[!t]
\centering
\small
\begin{tabular}{@{}lllll@{}}
\toprule
\% &
  Question ($q$) &
  Example 1 &
  Example 2 &
  Example 3 \\ \midrule
1 &
  \begin{tabular}[c]{@{}l@{}}what highschool did\\  harper lee go to?\end{tabular} &
  what did st augustine do? &
  who is keyshia cole dad? &
  \begin{tabular}[c]{@{}l@{}}when did charles goodyear\\ invented rubber?\end{tabular} \\
4 &
  \begin{tabular}[c]{@{}l@{}}what college did \\ albert einstein go to?\end{tabular} &
  \begin{tabular}[c]{@{}l@{}}what highschool did \\ harper lee go to?\end{tabular} &
  \begin{tabular}[c]{@{}l@{}}who did tim tebow play\\ college football for?\end{tabular} &
  what timezone is utah in? \\
40 &
  \begin{tabular}[c]{@{}l@{}}where did john mayer\\ go to college?\end{tabular} &
  \begin{tabular}[c]{@{}l@{}}what school did michael \\ jackson go to high school?\end{tabular} &
  \begin{tabular}[c]{@{}l@{}}where did derek fisher\\ go to college?\end{tabular} &
  \begin{tabular}[c]{@{}l@{}}what style of music does john\\ mayer play?\end{tabular} \\
90 &
  \begin{tabular}[c]{@{}l@{}}where did scott fitzgerald\\ go to college?\end{tabular} &
  who was f. scott fitzgerald? &
  \begin{tabular}[c]{@{}l@{}}where did otto frank\\ go to college?\end{tabular} &
  \begin{tabular}[c]{@{}l@{}}where did derek fisher go to\\ college?\end{tabular} \\ \bottomrule
\end{tabular}
\caption{Relevant examples fetched with time: as time proceeds, the examples fetched from the memory become increasingly relevant to the input question, leading to increasingly accurate predictions.}
\label{tab:webqarel}
\end{table*}

\begin{table*}[]
    \centering
    \small
    \begin{tabular}{|p{0.08\textwidth}|p{0.39\textwidth}|p{0.39\textwidth}|}
    \hline
         Timestep           &Question&                                               Neighbor        \\
    \hline
12 &    is not wanting a dirty person harassing you morally ok?   & is not wanting to hear foul language morally good? \\
70 &   when adam decided to stop living a lie and announces he's gay, are you expected to pretend you're straight when you're gay?     & are you expected to pretend you're straight when you're gay? \\
85  &  Should I help someone when they are having trouble?    & must you intervene if you see someone taking advantage of another disabled person? \\
230  & needing a hug to feel good.  &   is wanting to cuddle with loved ones okay? \\
\hline
    \end{tabular}
    \caption{Relevant examples retrieved at increasing timesteps: as time proceeds, the examples fetched from the memory become relevant to the input question, leading to accurate predictions.}
    \label{tab:neighbors-ert-cat}
\end{table*}

\paragraph{Inference} Let $k$ be the number of examples (\ie question-answer pairs) in the prompt.
For a given question $q$, We keep half~($k/2$) examples fixed in the prompt, whereas the other half $k/2$ examples are retrieved from a memory of feedback $M$.
As before, on receiving a question $q$, consults a memory $M$ to see if a similar question has been asked before.
However, different from earlier setups, in this case, we retrieve $k/2$ most similar questions from the memory $M$ on \textbf{which the system has been wrong earlier}.
The corresponding true answers are also retrieved.
These $k/2$ retrieved question-answer pairs are combined with the $k/2$ fixed questions to create a prompt, and query \gptshort.
Let $a'$ be the generated answer.

\paragraph{Growing memory of errors $M$} 
In our setup, we assume an expert user (or a teacher) that knows the true answer $a$ for a given query $q$.
The expert user compares the \gptshort generated answer $a'$ with $a$.
If the generated answer is correct ($a'=a$), no further action is taken.
If not, the entry ($(q, a)$) is added to the memory $M$.
As time passes, $M$ is populated with an increasing number of challenging examples that the model has been wrong on.
Thus, the retrieved $k/2$ examples get more relevant with time, aiding the accuracy.
In the experiments, we set $k=16$ due to budget constraints (note that the setups used in \citet{liu_what_2021} and \citet{Brown2020GPT3} set $k=64$, but their results are comparable to our baseline with $k=16$).

\begin{figure}[!h]
    \centering
    \includegraphics[width=\columnwidth]{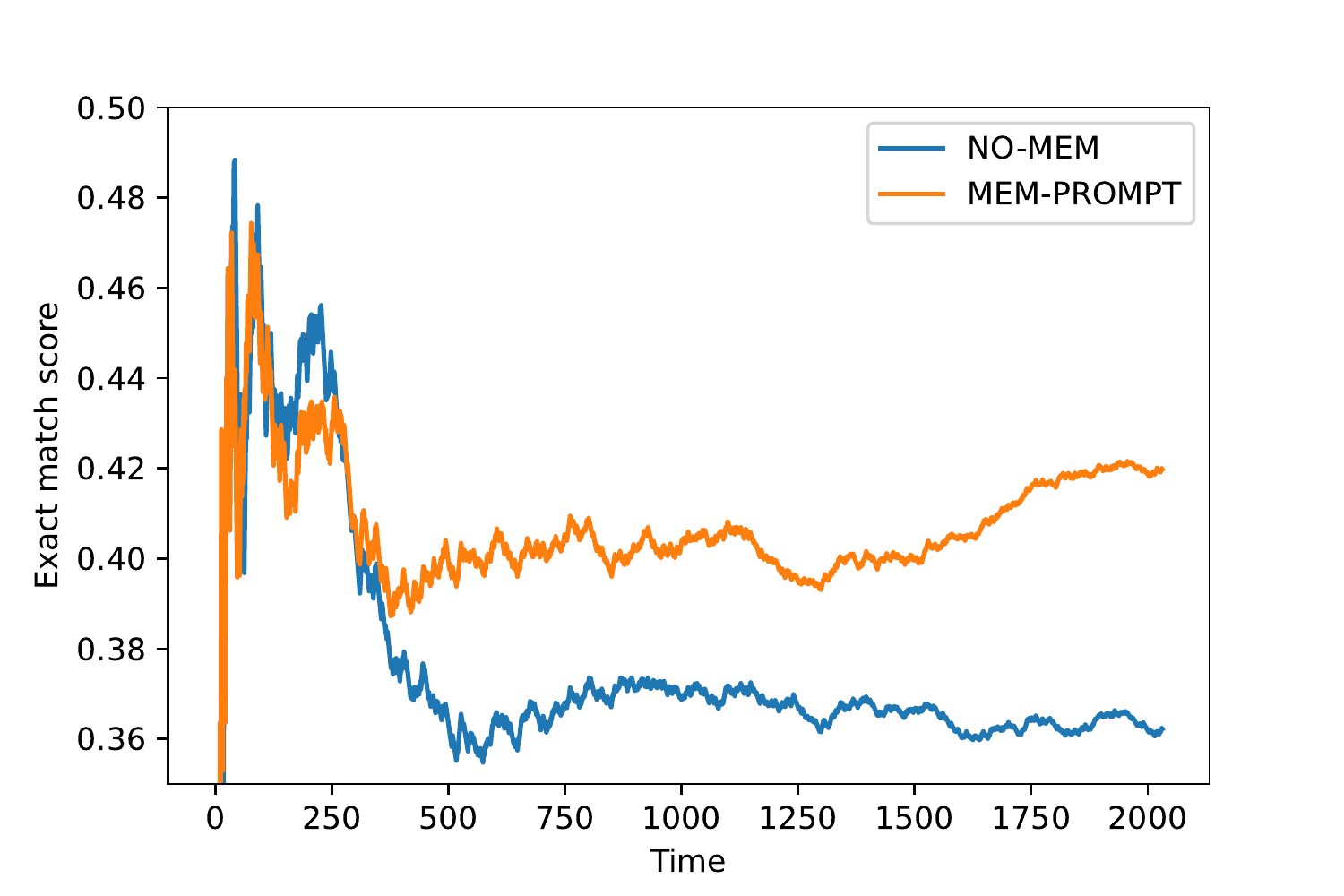}
    \caption{Instruction accuracy vs. time for \webqa.}
    \label{fig:webqaaccuracy}
\end{figure}

\paragraph{Results} Similar to \ert and word reasoning tasks, a memory of errors helps in increasing accuracy with time over 3,000 points in the test split of the \webqa dataset~(Figure~\ref{fig:webqaaccuracy}). This is expected, as $M$ gathers more examples on which \gpt has been wrong before. Adding these examples in the prompt avoids the model in repeating these mistakes.

To check if examples that belong to a similar domain improve with time, we cluster the questions in the test set of \webqa, and randomly select three clusters for our analysis.
Table~\ref{tab:webqarelcompletepart1} shows the top three of the 8 ($k=16/2$) examples retrieved from $M$ for the \textit{alma mater} cluster.\footnote{Additional examples are included in Appendix~\secref{sec:webqaappendix}.} All of these questions relate to the alma mater of famous personalities.
As the inference begins (with an empty $M$), the examples are not relevant to $q$. However, towards the end, almost all the samples are relevant to the given question.

\subsection{Factual question answering Examples}
\label{sec:webqaappendix}

Tables~\ref{tab:webqarelcompletepart1} and \ref{tab:webqarelcompletepart2} show additional examples for questions from \webqa which get additionally relevant examples as time proceeds.
The examples include questions that belong to the domains of Alma mater, Soccer, and Language. 
\begin{table*}[]
\centering
\begin{tabular}{@{}lrp{0.15\textwidth}p{0.15\textwidth}p{0.15\textwidth}p{0.15\textwidth}@{}}
\toprule
Domain &
  \multicolumn{1}{l}{\% Finished} &
  Question &
  Neighbor 1 &
  Neighbor 2 &
  Neighbor 3 \\ \midrule
Alma mater &
  1 &
  what highschool did harper lee go to? &
  what did st augustine do? &
  who is keyshia cole dad? &
  when did charles goodyear invented rubber? \\
Alma mater &
  5 &
  what college did albert einstein go to? &
  what highschool did harper lee go to? &
  who did tim tebow play college football for? &
  what timezone is utah in? \\
Alma mater &
  10 &
  what university did gordon brown attend? &
  what all does google now do?' &
  what team did david beckham play for in 2011?' &
  who did tim tebow play college football for?' \\
Alma mater &
  40 &
  where did john mayer go to college? &
  what school did michael jackson go to high school? &
  where did derek fisher go to college? &
  what style of music does john mayer play? \\
Alma mater &
  75 &
  where did john steinbeck go to college? &
  where did john mayer go to college? &
  what college did john stockton go to? &
 where did otto frank go to college? \\
Alma mater &
  95 &
  where did scott fitzgerald go to college? &
  who was f. scott fitzgerald? &
  where did otto frank go to college? &
  where did derek fisher go to college? \\ \midrule
Soccer &
  1 &
  what team did david beckham play for in 2011? &
  who did tim tebow play college football for? &
  what super bowl did peyton manning win? &
  what type of music did john lennon sing? \\
Soccer &
  25 &
  what team did ronaldo play for in 2003? &
  what part did winona ryder play in star trek? &
  what to do in richardson dallas? &
  who did the voice of darth vader in episode 3? \\
Soccer &
  33 &
  who did nasri play for before arsenal? &
  what year did ray allen join the nba? &
  who does donnie wahlberg play in the sixth sense? &
  what does david beckham play? \\
Soccer &
  65 &
  who has pudge rodriguez played for? &
  who does nolan ryan play for? &
  who did carlos boozer play for? &
  who does ronaldinho play for now 2011? \\
Soccer &
  99 &
  what team did david beckham play for before la galaxy? &
  who does david beckham play for? &
  what does david beckham play? &
  what team does david beckham play for in 2012? \\ \bottomrule
\end{tabular}
\caption{Relevant examples retrieved for \webqa \qa task~(Section~\secref{sec:webqaexperiments}). The retrieved examples get increasingly relevant as time proceeds.}
\label{tab:webqarelcompletepart1}
\end{table*}

\begin{table*}[]
\centering
\begin{tabular}{@{}lrp{0.15\textwidth}p{0.15\textwidth}p{0.15\textwidth}p{0.15\textwidth}@{}}
\toprule
Domain &
  \multicolumn{1}{l}{\% Finished} &
  Question &
  Neighbor 1 &
  Neighbor 2 &
  Neighbor 3 \\ \toprule
Language &
  1 &
  what does jamaican people speak? &
  when was ancient egypt created? &
  where is the denver broncos stadium located? &
  what is the name of the capital of spain? \\
Language &
  20 &
  what are the two official languages of paraguay? &
  what do portuguese people speak? &
  what language does cuba speak? &
  where is mission san buenaventura located? \\
Language &
  37 &
  what language does colombia? &
  what language does cuba speak? &
  what was the first language spoken in spain? &
  what is serbian language called? \\
Language &
  85 &
  what language does peru speak? &
  what are the official languages of the eu? &
  where is the latin language from? &
  what do portuguese people speak? \\
Language &
  90 &
  what language do they speak in colombia south america? &
  how many languages do they speak in spain? &
  where is the latin language from? &
  what language does cuba speak? \\ \bottomrule
\end{tabular}
\caption{Relevant examples retrieved for \webqa \qa task~(Section~\secref{sec:webqaexperiments}). The retrieved examples get increasingly relevant as time proceeds.}
\label{tab:webqarelcompletepart2}
\end{table*}


\section{Finding similar questions in low-resource settings}
\label{sec:lowresourceappendix}

\begin{figure}[!h]
    \centering
    \includegraphics[width=\columnwidth]{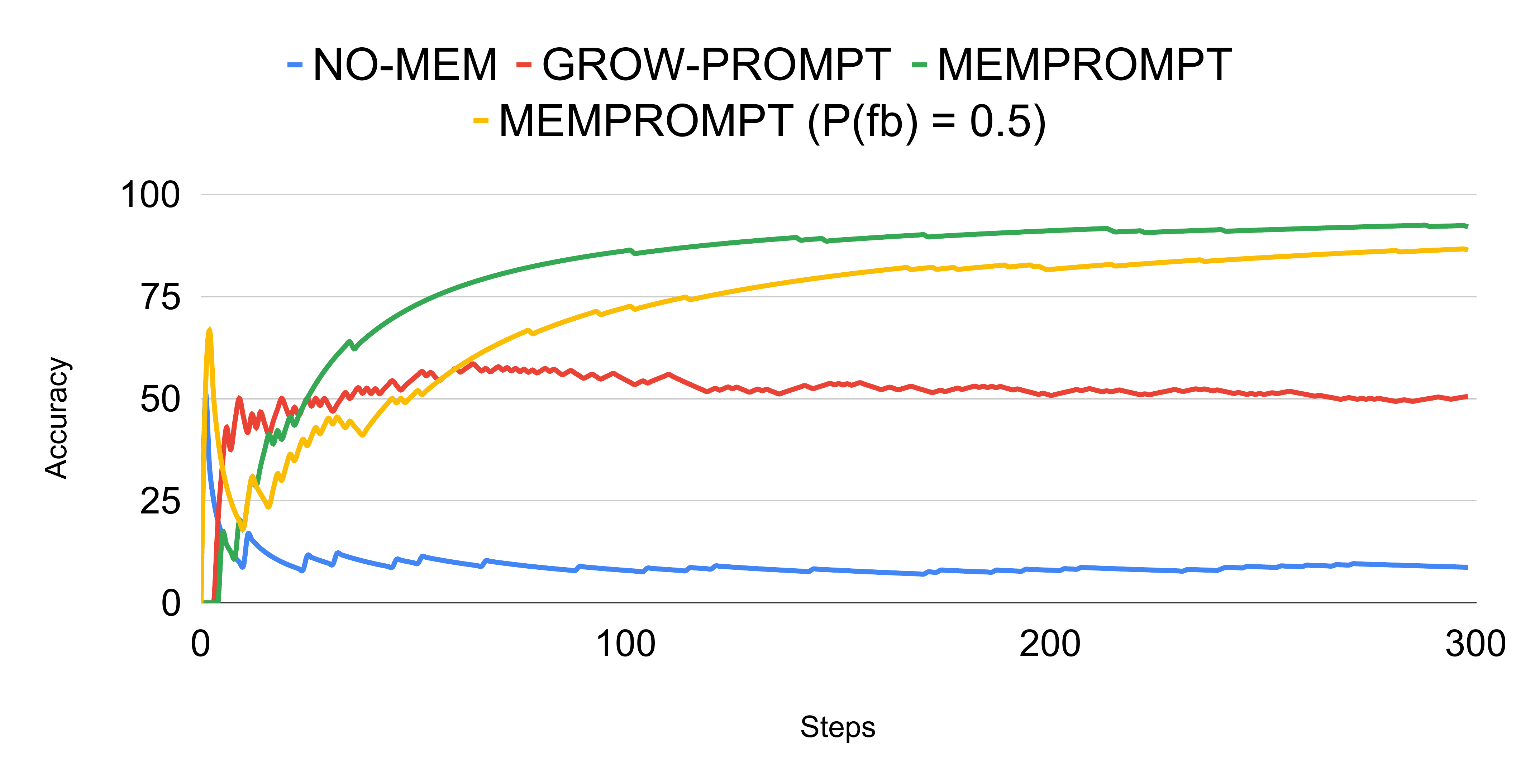}
    \caption{\textbf{Finding 2} Large gains on queries asked in English and Punjabi by \ours.}
    \label{fig:low-resource-gains}
\end{figure}

We also experimented using queries in Hindi and Punjabi, with
(English) feedback clarifying the queries' intent when GPT3 predictably misunderstands the task.Figure~\ref{fig:low-resource-gains} confirms significant gains using memory in this OOV setting. 
This setup highlights the case when the user does not speak fluent English and uses mixed language code, e.g.,  transcription in English and mixing words from another language to ask questions.

In low-resource settings~(\eg queries in transcribed Punjabi or Hindi), we perform similarity matching between a given question and a question in the memory by using surface-form similarity.
Specifically, we use Levenshtein distance to determine the closest query in the memory.
We note that as the memory grows large, we can use mechanisms such as FAISS~\citep{johnson2019billion} for trained memory, and suffix-trees for fast retrieval using surface form similarity.

\section{Sample results}
Table~\ref{tab:wrongwithoutmem} shows randomly sampled \quesm-\ansm pairs, and the corresponding \ansm generated by \gpt and \ours.
The complete set of outputs is located in the project repository~\url{https://github.com/madaan/memprompt/tree/main/results}.
\newpage
\clearpage

\lstset{basicstyle=\small\ttfamily,columns=fullflexible}
\begin{lstlisting}[linewidth=0.95\linewidth, xleftmargin=.1\textwidth, breaklines=true,language=Python,float=*, label=code1, caption=Sample templates for the five tasks.]
templates = [
        {
            "type": "syn",
            "template_id": "syn1",
            "question": lambda word1: f"What is similar to < {word1} > ?",
            "question_clarification": lambda word1: f"What is similar to < {word1} > ? | clarification: when I ask for similar to , I want a synonym.",
            "clarification": "clarification: when I ask for similar to , I want a synonym.",
            "answer": lambda word1, word2: f"the synonym for {word1} is {word2}",
        },
        {
            "type": "ant",
            "template_id": "ant0",
            "question": lambda word1: f"What is unlike < {word1} > ?",
            "question_clarification": lambda word1: f"What is unlike < {word1} > ? | clarification: when I ask for unlike , I want an antonym.",
            "clarification": "clarification: when I ask for unlike , I want an antonym.",
            "answer": lambda word1, word2: f"the antonym for {word1} is {word2}",
        },
        {
            "type": "defn",
            "template_id": "defn0",
            "question": lambda word: f"< {word} > means what ?",
            "question_clarification": lambda word: f"< {word} > means what ? | clarification: when I ask for means what , I want a definition.",
            "clarification": "clarification: when I ask for means what , I want a definition.",
            "answer": lambda word, definition: f"the definition of {word} is {definition}",
        },
        {
            "type": "sent",
            "template_id": "sent1",
            "question": lambda word: f"< {word} > can be used how ?",
            "question_clarification": lambda word: f"< {word} > can be used how ? | clarification: when I ask for can be used how , I want a sentence.",
            "clarification": "clarification: when I ask for can be used how , I want a sentence.",
            "answer": lambda word, sentence: f"a sentence with {word} is: {sentence}",
        }]
\end{lstlisting}

\newcommand{\tabwidthsampletask}{0.4}
\begin{table*}[] 
 \centering
\begin{tabular}{p{\tabwidthsampletask\textwidth}p{\tabwidthsampletask\textwidth}p{0.1\textwidth}}
\toprule
\textbf{Question~(\quesm)} & \textbf{Answer~(\ansm)} &  type \\
\midrule
                  What is the opposite of < misconstrue > ? &                                                         the antonym for misconstrue is verify &  \ant \\
                        What is the opposite of < gross > ? &                                                               the antonym for gross is polite &  \ant  \\
                                  expand on < chelicera > ? &                       the definition of chelicera is One of the anterior pair of mouth organs & \defn \\
what is a sentence that can be used to define < mawseed > ? &                                     the definition of mawseed is The seed of the opium poppy. & \defn \\
                      what has a < bitt > like ring to it ? &                                                                   the homophone for bitt is bit &  \homn \\
                      what can one confuse with < holed > ? &                                                                 the homophone for holed is hold &  \homn \\
                               < spread > can be used how ? &                        a sentence with spread is: a couple of sheep are spread out in a field & \sent \\
                              make something with < pot > ? &                                    a sentence with pot is: bonsai tree in pot at zen garden . & \sent \\
                             What is akin to < musician > ? &                                                   the synonym for musician is instrumentalist &  \syn \\
                               What is akin to < zigzag > ? &                                                                the synonym for zigzag is move &  \syn \\
\bottomrule
\end{tabular}

\caption{Sample \quesm-\ansm pairs in English. The same type of question can be asked in multiple ways. Our method makes no assumptions as to how a question might be asked.}
\label{tab:linguistictasks} 
\end{table*}

\begin{table*}[] 
 \centering
\begin{tabular}{p{\tabwidthsampletask\textwidth}p{\tabwidthsampletask\textwidth}p{0.1\textwidth}}
\toprule
\textbf{Question~(\quesm)} & \textbf{Answer~(\ansm)} &  type \\
\midrule
           < tabulate > ka ulta kya hai ? &                                                                           the antonym for tabulate is randomize &  \ant \\
              < foot > ka vilom kya hai ? &                                                                                    the antonym for foot is head &  \ant \\ 
     < lettish > ka matlab kya hota hai ? &                                      the definition of lettish is The language spoken by the Letts. See Lettic. & \defn \\
            < housing > ka arth kya hai ? &                                  the definition of housing is An appendage to the hames or collar of a harness. & \defn \\
       sunne mai < perl > jaisa kya hai ? &                                                                                   the homophone for perl is pearl &  \homn \\ 
   < council > jaisa kya sunai deta hai ? &                                                                                the homophone for council is conceal &  \homn \\
  < city > ko ek vakya mai kaise likhen ? &                            a sentence with city is: the city takes on an even more interesting hue during event & \sent \\
   < fly > ko ek vakya mai kaise likhen ? &                                                         a sentence with fly is: airplane fly into a storm cloud & \sent \\
   < critique > kai samaan kya hota hai ? &                                                                            the synonym for critique is evaluate &  \syn \\
< psychiatric > kai samaan kya hota hai ? &                                                                           the synonym for psychiatric is mental &  \syn \\
\bottomrule
\end{tabular}

 \caption{Sample \quesm-\ansm pairs in Hindi.}
\label{tab:hinditasks} 

\begin{tabular}{p{\tabwidthsampletask\textwidth}p{\tabwidthsampletask\textwidth}p{0.1\textwidth}}
\toprule
\textbf{Question~(\quesm)} & \textbf{Answer~(\ansm)} &  type \\
\midrule
             < edit > de ult ki hunda ae ? &                                                                            the antonym for edit is expand &  \ant \\
               < spring > ton bhin ki ae ? &                                                                        the antonym for spring is compress &  \ant \\
       < patricide > di paribhasha dasso ? &                the definition of patricide is The crime of one who murders his father. Same as Parricide. & \defn \\
       < sankhya > da matlab ki hunda ae ? & the definition of sankhya is A Hindoo system of philosophy which refers all things to soul and a rootless germ called prakriti & \defn \\
        sunnan vich < site > varga ki ae ? &                                                                             the homophone for site is sites &  \homn \\
< chance > da samnam ki ae ? &                                                                               the homophone for chance is chants &  \homn \\
  < enter > nu ek vak vich kidan vartiye ? &                       a sentence with enter is: as you enter you will be greeted by these rows of trees . & \sent \\
                < tree > da prayog ki ae ? &                              a sentence with tree is: wide shot of an eagle and its nest on top of a tree & \sent \\
          < unlawful > jidan ki hunda ae ? &                                                                       the synonym for unlawful is illegal &  \syn \\
         < constrict > jidan ki hunda ae ? &                                                                        the synonym for constrict is close &  \syn \\
\bottomrule
\end{tabular}

\label{tab:punjabitasks} 
 \caption{Sample \quesm-\ansm pairs in Punjabi.}

\end{table*}

\newcommand{\tabwidthsample}{0.18}

\begin{table*}[] 
 \centering
\begin{tabular}{p{\tabwidthsample\textwidth}p{\tabwidthsample\textwidth}p{\tabwidthsample\textwidth}p{\tabwidthsample\textwidth}}
\toprule
                                                   \textbf{Question}~(\quesm) &                                                                                                                                                           \textbf{Expected answer}~(\ansm$^*$) &                                                                                                                         \textbf{\gptshort generated answer} &                                                         \textbf{\ours generated answer} \\
\midrule
            give me something i would hear as < whether > ? &                                                                                                                             the homophone for whether is whither &                                   the example of something you would hear as whether is a question word used to ask about possibilities. &                                                    the homophone for whether is weather \\ \hline
                      what has a < mane > like ring to it ? &                                                                                                                                   the homophone for mane is made &                                                                                         the word that has a mane like ring to it is mane &                                                          the homophone for mane is main \\ \hline
                                 What is like < confine > ? &                                                                                                                                the synonym for confine is hold &                                                                                                      the homophone for confine is confined &                                                   the synonym for confine is restrain \\ \hline
                                     < leper > means what ? &                                                                                                     the definition of leper is a person affected with leprosy. &                                        a sentence with leper is: a leper is a person with a disease that makes his skin scaly and white. &                                  the definition of leper is a person who has leprosy. \\ \hline
                                   expand on < chaperon > ? &                                                                                  the definition of chaperon is a matron who accompanies a young lady in public &                                                                                          the expansion of chaperon is one who chaperons. & the definition of chaperon is a woman who goes with a young girl to social functions. \\ \hline
what is a sentence that can be used to define < wassail > ? &                                                                        the definition of wassail is an ancient expression of good wishes on a festive occasion &                                                                    a sentence with wassail is: < wassail > wassails the christmas season &                               the definition of wassail is to drink to the health of. \\ \hline
                                 expand on < paralbumin > ? & the definition of paralbumin is a proteidlike body found in the fluid from ovarian cysts and elsewhere. it is generally associated with a substance related to & the expansion of paralbumin is an albuminoid globulin occurring in the blood of the horse and other animals, and in the urine of humans. &                         the definition of paralbumin is a particular kind of albumin. \\ \hline
                        what can one confuse with < kew > ? &                                                                                                                                     the homophone for kew is key &                                                                                              a word that can be confused with kew is due &                                                            the homophone for kew is cue \\ \hline
                               What is like < pneumatic > ? &                                                                                                                             the synonym for pneumatic is windy &                                                                                                   the homophone for pneumatic is hydraulic &                                                     the synonym for pneumatic is airy \\ \hline
             give me something i would hear as < seeder > ? &                                                                                                                                 the homophone for seeder is seer &                            something you would hear as seeder is: someone who plants things in the ground or causes seeds to be planted. &                                                        the homophone for seeder is said \\ 
\bottomrule
\end{tabular}

 \caption{A subset of random samples where \gpt without memory was incorrect.}
\label{tab:wrongwithoutmem} 
\end{table*}

\end{document}